\documentclass[conference]{IEEEtran}
\IEEEoverridecommandlockouts
\usepackage{cite}
\usepackage{amsmath,amssymb,amsfonts}
\usepackage{algorithmic}
\usepackage{graphicx}
\usepackage{textcomp}
\usepackage{xcolor}
\usepackage{times}  
\usepackage{helvet}  
\usepackage{courier}  
\usepackage[hyphens]{url}  
\usepackage{graphicx} 
\usepackage{caption} 
\usepackage{epsfig}
\usepackage{array}
\usepackage{comment}
\usepackage{multirow}
\usepackage{subcaption}
\usepackage{hhline}
\usepackage{algorithm}
\usepackage{caption}
\usepackage{booktabs}

\def\BibTeX{{\rm B\kern-.05em{\sc i\kern-.025em b}\kern-.08em
    T\kern-.1667em\lower.7ex\hbox{E}\kern-.125emX}}
\begin{document}

\title{Mitigating Data Absence in Federated Learning Using Privacy-Controllable Data Digests}

\author{\IEEEauthorblockN{Chih-Fan Hsu}
\IEEEauthorblockA{\textit{Digital Center} \\
\textit{Inventec Corporation}\\
Taipei, Taiwan \\
hsu.chih-fan@inventec.com}
\and
\IEEEauthorblockN{Ming-Ching Chang}
\IEEEauthorblockA{\textit{Department of Computer Science} \\
\textit{University at Albany}\\
Albany, NY, United States \\
mchang2@albany.edu}
\and
\IEEEauthorblockN{Wei-Chao Chen}
\IEEEauthorblockA{\textit{Digital Center} \\
\textit{Inventec Corporation}\\
Taipei, Taiwan \\
chen.wei-chao@inventec.com}
}

\maketitle

\begin{abstract}
The absence of training data and their distribution changes in federated learning (FL) can significantly undermine model performance, especially in cross-silo scenarios. To address this challenge, we introduce the Federated Learning with Data Digest (FedDig) framework. FedDig manages unexpected distribution changes using a novel privacy-controllable data digest representation. This framework allows FL users to adjust the protection levels of the digest by manipulating hyperparameters that control the mixing of multiple low-dimensional features and applying differential privacy perturbation to these mixed features. Evaluation of FedDig across four diverse public datasets shows that it consistently outperforms five baseline algorithms by substantial margins in various data absence scenarios. We also thoroughly explored FedDig's hyperparameters, demonstrating its adaptability. Notably, the FedDig plugin design is inherently extensible and compatible with existing FL algorithms.
\end{abstract}


\section{Introduction}
\label{sec:intro}

\IEEEPARstart{F}{ederated} Learning (FL) has shown great success in addressing privacy and data-sharing concerns in distributed learning~\cite{FML:Survey:CST2021}. However, practical challenges emerge in FL when competition or shifts in collaboration intentions cause clients to leave or become unavailable, resulting in extended {\it data absence}.
The data absence issue is typically minor in cross-device FL due to data similarity among clients, as available clients can compensate for the impact. Data absence presents a significant challenge for the cross-silo FL setting, as the lack of data providers can severely undermine FL training efficacy, especially when data distributions between clients are non-independent and identically distributed (non-IID)~\cite{FL:nonIID:arXiv2021}. 

Existing FL methods can only tolerate short, temporary client departures~\cite{Advances:Open:Problems:FL:2021}. The cross-silo FL setting becomes particularly prominent in real-world scenarios, where collaborative entities, such as hospitals or companies, are both scarce and possess distinct datasets. Consequently, the applicability of existing FL algorithms is limited.

\begin{figure*} [t]
\centerline{
  \begin{subfigure}[b]{0.53\linewidth}
  \centerline{
  \includegraphics[width=\linewidth]{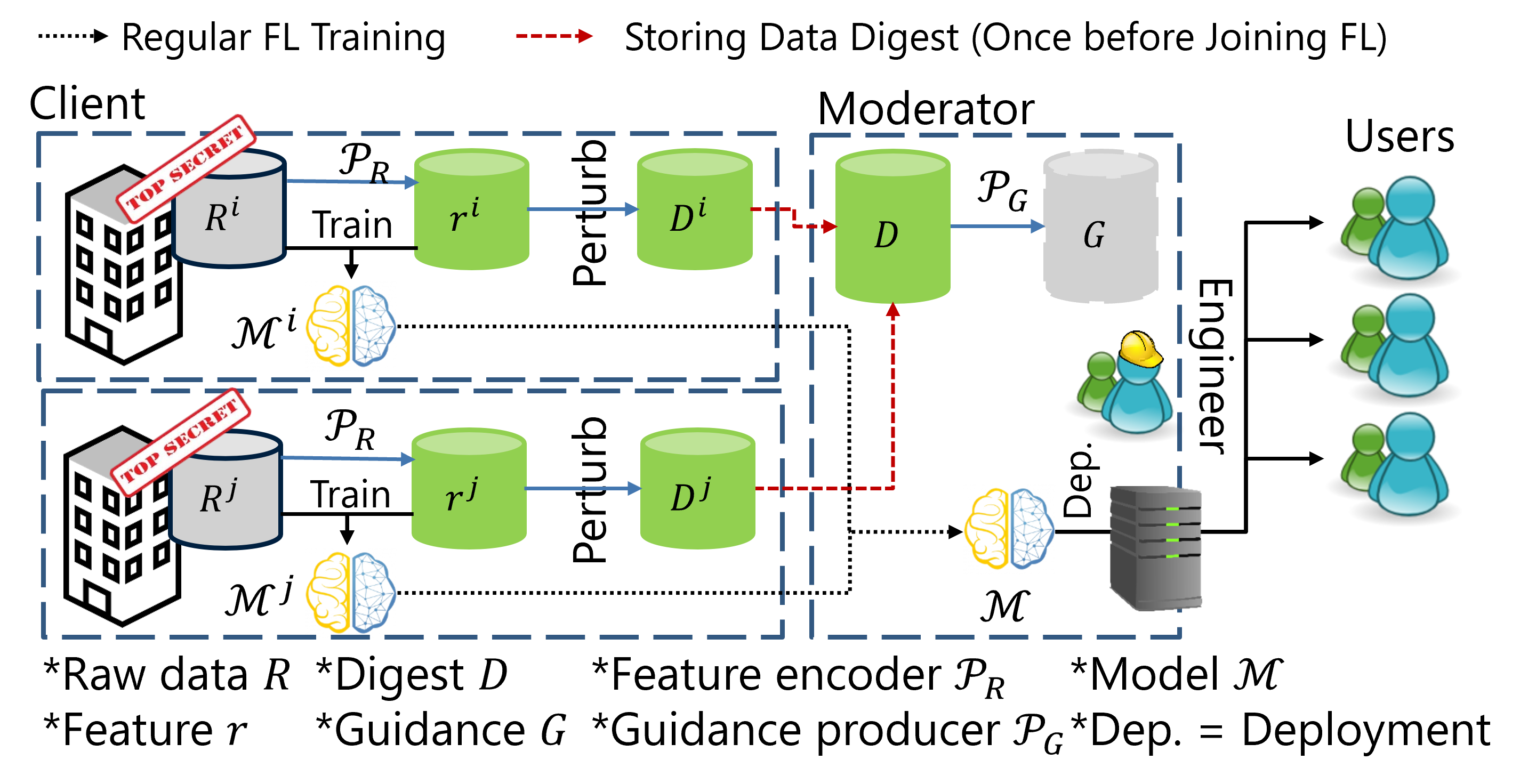}
  }
  \label{fig:Collecting}
  \caption{FL training with clients available.}
  \end{subfigure}
  \hspace{3mm}
  \begin{subfigure}[b]{0.47\linewidth}
  \centerline{
  \includegraphics[width=\linewidth]{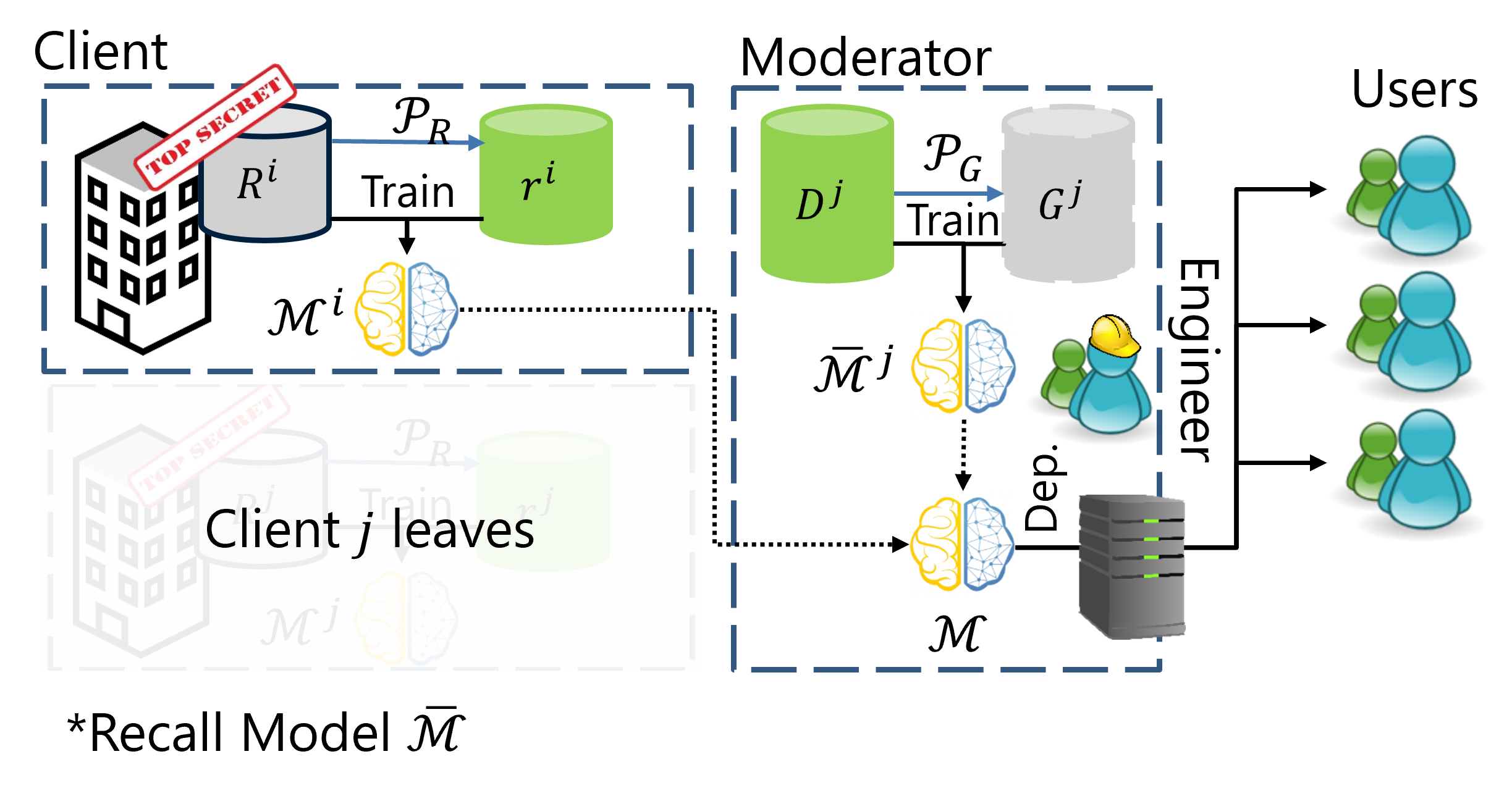}
  }
  \label{fig:Training}
  \caption{FL training with client absence.}
  \end{subfigure}
}
\caption{The FedDig framework overview: 
(a) For an available client $i$, data digest $D^i$ is generated and stored at the moderator before joining the FL training. The available clients behave like a regular federated learning algorithm, such as FedAvg~\cite{FedAVg:AISTATS2017}.
(b) When a client $j$ is absent, the moderator synthesizes the model update $\nabla {\cal \overline{M}}^j_t$ of the client $j$'s data using a recalled model ${\cal \overline{M}}$ and the digest ${\cal D}^j$. This approach addresses training data distribution change due to client absence.
}
\label{fig:Stages}
\end{figure*}

Relying solely on memorizing historical gradients to address data absence is ineffective, as these gradients gradually lose their representativeness during training. Instead, we propose a more effective approach, wherein we hypothesize that {\it synthesizing gradient updates for the missing data during FL training can mitigate the impact of data absence}. By incorporating a mechanism for memorizing and sharing another form of data, we can generate synthetic gradients to improve FL training. 

A key challenge in this approach is ensuring that the data sharing mechanism complies with the privacy standard. The definition of privacy can vary depending on the users or tasks involved. For instance, in industries, privacy considerations often hinge on legal judgments, such as determining whether a breach of business secrets has occurred. A comprehensive privacy definition across disciplines is still lacking. Therefore, our goal is to seek a method that allows FL users to manage and control the privacy level of shared data.

We introduce a framework called {\bf Federated Learning using Data Digests (FedDig)} to address data absence by synthesizing gradients for missing data at the FL moderator. 
This innovative approach utilizes a privacy-controllable data representation known as the {\bf data digest}. The FedDig framework, illustrated in Figure~\ref{fig:Stages}, includes two main components: (1) a method for generating data digests with controllable privacy levels and (2) a technique to synthesize gradients for missing data based on these digests. 

To create data digests with controllable privacy levels, we first encode raw data into low-dimensional features. These features are then perturbed using two protection strategies: feature mixing~\cite{Survey:Img:Mix:Data:Aug} and a noise-adding mechanism based on Differential Privacy (DP)~\cite{DP2006}. 
The feature mixing design makes it exceptionally difficult to reconstruct raw samples from the digests, while the DP method provides mathematical guarantees that the sample identity is obscured. Despite the evolving nature of data privacy legislation and application conditions~\cite{Privacy:FL:GDPR}, FedDig allows users to control the privacy level of the shared data digest via hyperparameters. This allows users to balance privacy protection with model performance.

FedDig is an end-to-end training framework that seamlessly integrates with existing FL algorithms and can readily accommodate new FL algorithms with similar gradient aggregation architectures. To evaluate the performance of FedDig, we devised four typical data absence scenarios. The results demonstrate that FedDig effectively mitigates the loss of accuracy in all tested scenarios. We also conducted a comparative analysis between FedDig and five widely used FL algorithms—FedAvg, FedProx, FedNova, FedDyn, and AdaBest—using four public datasets (EMNIST, Fashion-MNIST, CIFAR-10, and CIFAR-100). This evaluation includes challenging scenarios, including clients sequentially departing FL training without rejoining. The experimental results highlight the robust efficacy of FedDig in addressing the challenges associated with data absence. The contributions of this paper are summarized in the following:
\begin{itemize}

\item We propose the FedDig framework to address client data absence issues in FL.

\item We design the data digest as a privacy-controllable data representation that can synthesize the missing data gradients for FL training. We also provide a quantitative assessment to evaluate the privacy risk of the data digest.

\item We conduct extensive experiments to evaluate FedDig using various hyperparameters. Across four public datasets and scenarios involving client absence, FedDig consistently outperforms five prevalent FL algorithms. Additionally, FedDig is adaptable and can enhance the performance of existing algorithms.

\end{itemize}

\section{The Impact of Data Absence in Cross-Silo FL}
\label{sec:pre}

We demonstrate the influence of client data absence on FL in a cross-silo setting with four clients, using the CIFAR-10 dataset.
To simulate this scenario, we distribute the original training data among four clients (C1 to C4) using the Dirichlet distribution~\cite{Li2021PracticalOF} with a hyperparameter $\mu = 0.1$, creating a highly skewed and non-IID setting.
The data for each client are divided into three subsets with 80\%, 10\%, and 10\% for training, validation, and testing, respectively.
The FL moderator retains the original CIFAR-10 test set to monitor the overall test accuracy.

\begin{figure}[t]
\centerline{
  \begin{subfigure}[b]{=0.5\linewidth}
  \centerline{
  \includegraphics[width=\linewidth]{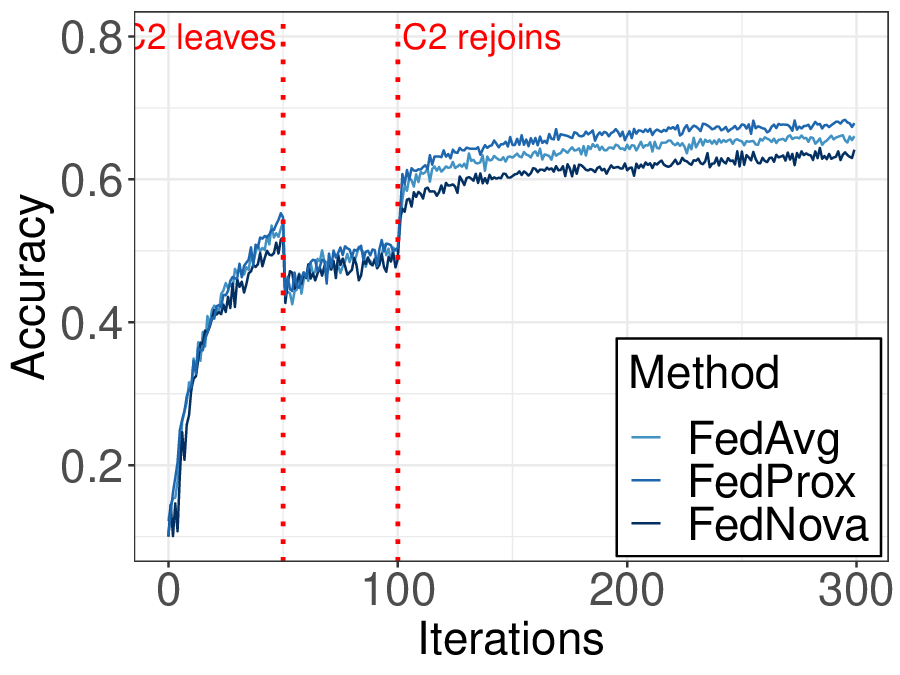}
  }
  \caption{Overall test accuracy}\label{fig:OTA}
  \end{subfigure}
  \hspace{0.3cm}
  \begin{subfigure}[b]{=0.5\linewidth}
  \centerline{
  \includegraphics[width=\linewidth]{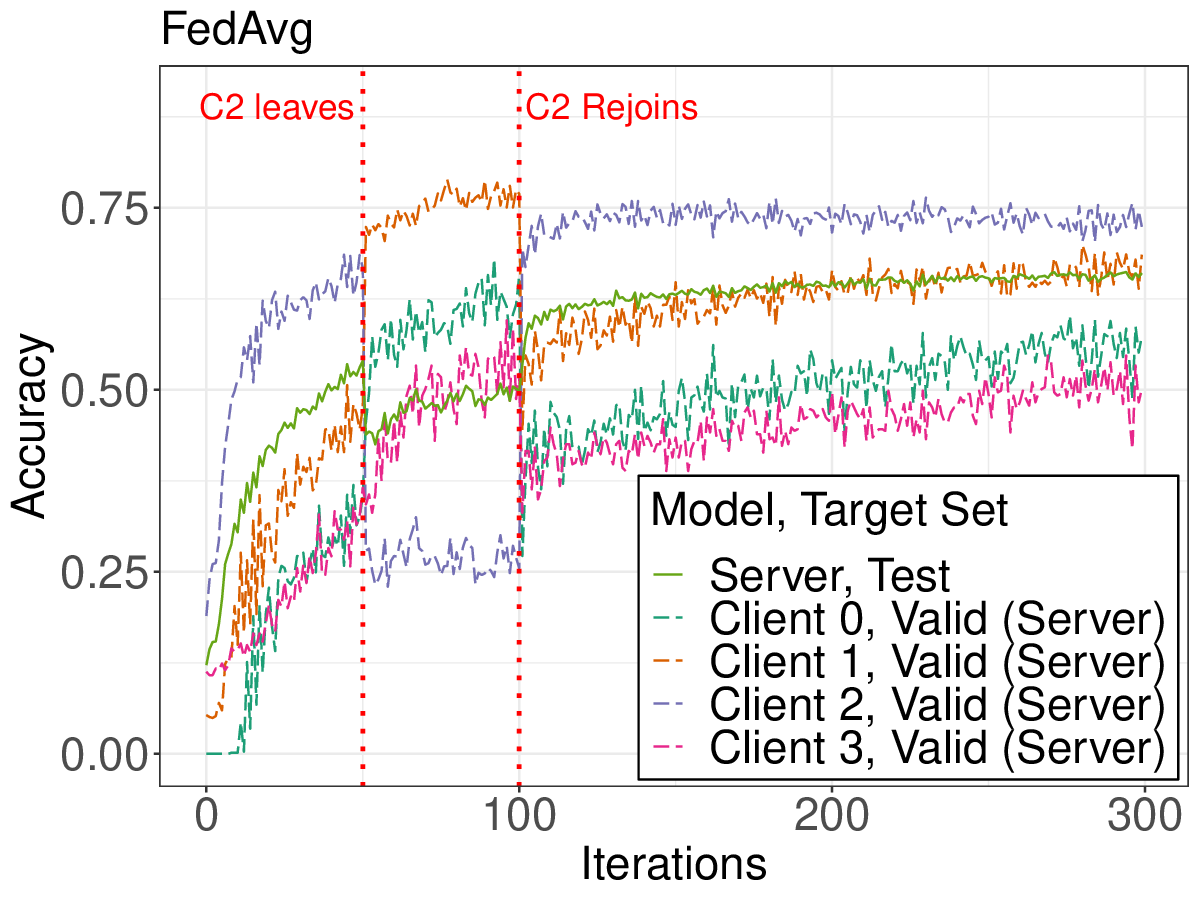}
  }
  \caption{Client validation accuracy}\label{fig:CVA}
  \end{subfigure}
}
\caption{Test accuracy drops caused by client/data absence.}
\label{fig:data-absent}   
\end{figure}

Throughout FL training, we simulate the departure of the largest client (C2), which holds the most extensive data, at iteration 50, with a rejoining at iteration 100. Figure~\ref{fig:OTA} shows the test accuracy on the moderator test set for three common FL algorithms. A noticeable drop in accuracy follows the client's departure, indicating that the server model quickly forgets learned knowledge after a few updates.

Figure~\ref{fig:CVA} displays the validation accuracy of the clients in the case of FedAvg. The model struggles to maintain C2's data distribution after its departure. Although the accuracy increases for the remaining clients, likely due to the reduced complexity of the classification problem, the overall test accuracy declines.

\section{Related Works}
\label{sec:related}

Federated Learning (FL) methods are designed to address data security and privacy issues in distributed, collaborative learning~\cite{Survey_Security_Privacy_FL_FGCS2020}.
An FL server aggregates model updates (such as gradients) from individual clients and merges them, mitigating various data privacy and security concerns. Effective model aggregation for FL training with non-IID client data distribution has been extensively studied.
FedAvg~\cite{FedAVg:AISTATS2017} is a simple, widely used algorithm that aggregates local client updates via weighted averaging.
However, non-IID data pose challenges for FedAvg, leading to inferior performance and complicating convergence analysis~\cite{FedAvg:Convergence:ICLR2020}.
FedProx~\cite{FedProx:MLSys2020} addresses these issues by adding an $\ell_2$ regularization term with a new hyper-parameter 
to restrict aggressive local client updates.
FedNova~\cite{FedNova:NeurIPS2020} modifies the aggregation of FedAvg by normalizing updates relative to the number of training steps each client takes. The Scaffold method~\cite{SCAFFOLD:PMLR2020} estimates the gradient update direction and uses it to adjust the direction of local updates to prevent client drifts. FSL~\cite{FSL:TAI2024} leverages small amount of more representative data on the server to correct gradient updates and mitigate the drift caused by non-IID data.
FedDyn~\cite{FedDyn:ICLR2021} and AdaBest~\cite{AdaBest} introduce a regularization term in local training based on a global model and the local model from the previous iteration.
FedBN~\cite{FedBN:ICLR2021} updates Batch Normalization (BN) layers locally without uploading them to the moderator.
In~\cite{RatioLoss_NonIID_AAAI2021}, a monitoring scheme infers the composition of training data in each FL training round, mitigating data imbalance effects using Ratio Loss.
In~\cite{MOON:CVPR2021}, local training are corrected by comparing representations learned by global and local models in current and previous rounds.
The experience-driven Q-learning of FAVOR~\cite{FAVOR:INFOCOM2020} intelligently controls client participation in FL rounds, counterbalancing biases introduced by non-IID data. While these methods address data imbalance problems, none specifically focus on the critical issues of client data absence.

Making the data distribution IID between clients by sharing information about client data using techniques like image mixing, data distillation, and augmentation~\cite{Survey:Img:Mix:Data:Aug} could potentially mitigate the issues of the client absence. Some notable works in this area include InstaHide~\cite{InstaHide}, DataMix~\cite{DataMix}, XORMixUp~\cite{XORMixUp}, and FedMix~\cite{yoon2021fedmix}. 
XORMixUp works by collecting encoded samples from other clients and decoding them to generate synthetic samples, creating an IID dataset for model training. The encoding and decoding processes use bit-wise XOR to ensure data privacy. The FedMix framework employs a mixup technique, merging multiple raw training data at each client. Clients then send and receive mixed raw data to and from other participants. However, sending averaged data can lead to privacy leakage, and the authors did not provide a thorough privacy definition or discuss related privacy concerns.


\medskip
\noindent
{\bf Client absence in FL:} 
Methods reviewed in~\cite{FL:Unreliable:Clients:arXiv2021} identify client departures and adaptively re-weight client gradients during aggregation.
DeepSA~\cite{FL:Unreliable:Clients:arXiv2021} handles adversarial clients using secure aggregation.
FedCCEA~\cite{FedCCEA:arXiv2021} estimates client contributions based on data quality and excludes the least contributive clients from test estimation.
Continual Federated Learning (CFL) is designed to mitigate catastrophic forgetting in FL scenarios. CFeD~\cite{ijcai2022p303} uses knowledge distillation with additional unlabeled surrogate datasets on both the server and client sides to revisit knowledge related to unavailable data. However, this method requires clear task separation, and detecting tasks in CFL remains an open question.
ECFL~\cite{Casado2023} employs an ensemble structure, where the server maintains a set of client models with a selection mechanism, though determining the optimal number of stored models is challenging.
GradMA~\cite{Luo_2023_CVPR} uses quadratic programming to extract knowledge from pre-stored client updates at the server, but choosing an appropriate client update is difficult as gradients can become unrepresentative after several updates.
To our knowledge, none of these studies addresses the issue of long-term client absence in FL, raising practical concerns that have not been  adequately addressed.






\section{The FedDig Framework}
\label{sec:method}

The FedDig framework aims to mitigate the impact of data absence in Federated Learning (FL). It enables the FL moderator to continue training as if all clients were present by using synthetic gradients generated from privacy-controllable data digests. This section also describes the privacy assessment for the data digests.

\subsection{The FedDig Training Framework}
\label{sec:training}

FedDig enhances the standard FedAvg framework~\cite{FedAVg:AISTATS2017}, by introducing a novel design of data digest generation and transmission, which incurs only a small computation overhead, and includes an additional gradient synthetic process at the FL server.
Specifically, the FedDig includes three key components: (1) Data digest $D$: This is generated at each client using a fixed {\em feature encoder} ${\cal P}_R$. A feature perturbing process controls the privacy level of the data digest. (2) Training guidance $G$: This is generated from the data digest at the server using a trainable {\em guidance producer} ${\cal P}_G$. When the moderator detects any client absence, a recall model ${\cal \overline{M}}$ can be trained with $D$ and $G$ to produce synthetic gradients for missing data. (3) An additional training process jointly updates ${\cal P}_G$ and the server model $\cal M$. It is designed to automatically obtain suitable training guidance for synthesizing representative gradients. 


Figure~\ref{fig:Stages} illustrates two key scenarios in the FedDig training framework: (a) the generation and collection of digests when clients are available, and (b) the synthesis of gradients from these digests when any client becomes absent. The detailed training process of FedDig is outlined in Algorithm~\ref{alg:FedSyn_psudocode}.

\medskip
\noindent
{\bf Training Losses:}
For each available client $i$, client model ${\cal M}^i$ is trained using raw data $R^i$ and encoded features $r^i$ produced by a data encoder ${\cal P}_R$ using the standard cross-entropy loss:
${\cal L}_{client}^{avail} = {\cal L}_{ce}
\left(
{\cal M}^i (R^i, r^i), y^i
\right).$
For each missing client $j$, the FL moderator synthesizes the gradient of the missing data by training a recall model ${\cal \overline{M}}^j$ with the guidance $G^j$ and the pre-stored digests $D^j$ using cross-entropy loss:
\begin{equation}
{\cal L}_{client}^{absent} = {\cal L}_{ce}
\left(
{\cal \overline{M}}^j \left({\cal P}_G(D_R^j), D_R^j \right), D_y^j
\right),
\end{equation}
where $G^j$ is produced from $D^j$ by a trainable guidance producer ${\cal P}_G$. The digest $D$ contains a set of mixed and perturbed low-dimensional features $D_R$ and their corresponding mixed soft labels $D_y$. 

The moderator is expected to generate suitable $G^j$ from $D^j$ using ${\cal P}_G$ to solve a multi-class classification problem.
To achieve this, the server model is updated with the following loss function after aggregating all client models: 
%
\begin{equation}
{\cal L}_{server} = {\cal L}_{ce} 
\left(
{\cal A} ({\cal P}_G(D_R), D_R), D_y
\right),
\end{equation}
where the aggregated model ${\cal A}$ represents the updated server model after incorporating both client and synthetic updates. Specifically, 
${\cal A} = {\cal M} + \sum_i{a^i}{\nabla \cal M}^i + \sum_j{{a^j}{\nabla \cal \overline{M}}^j}$,
where $a$ represents the aggregation weight, and $\sum_i{a^i} + \sum_j{a^j} = 1$. We set all aggregation weights to $1/n$, with $a^i = a^j = 1/n$, where $n$ is the total number of participating clients.
%
%

%
%



\begin{algorithm}[t]
\caption{FedDig Training} 
\label{alg:FedSyn_psudocode}
\begin{algorithmic}[1]
    \STATE \textbf{Initialize}: ${\cal M}$ and ${\cal P}_G$\\

    \FOR{each FL client $c=1, \ldots, n$ {\bf in parallel}}
        \STATE Client $c$ produces digests $D^i=(D^i_R,D^i_y)$ and pushes $D^i$ to the moderator.
    \ENDFOR
    \FOR {each training iteration $t$}
        \STATE Moderator pushes server model ${\cal M}_t$ to all clients.
        \STATE \textcolor{blue}{// Client-side training}
        \FOR {available client $i=1, \ldots, m$ {\bf in parallel}}
            \STATE Client $i$ generates $\nabla {\cal M}^i_t$ with the loss ${\cal L}_{client}^{avail}$ and pushes model gradient ${\nabla \cal M}^i_t$ to the moderator.
        \ENDFOR
        \STATE \textcolor{blue}{// Moderator-side training (parallel to the client-side training)}
        \FOR {absent client $j=1, \ldots, k$ {\bf in parallel}}
            \IF{the set of digests $D^j$ exists}
                \STATE Generate recall model ${\cal \overline{M}}^j_t$ from ${\cal {M}}_t$.
                \STATE Synthesize $\nabla {\cal \overline{M}}^j_t$ using loss ${\cal L}_{client}^{absent}$.
            \ENDIF
        \ENDFOR
        \STATE Moderator updates ${\cal A}_t$ using ${\cal M}_t$, $\nabla {\cal M}^i_t$, and $\nabla {\cal \overline{M}}^j_t$.
        \STATE \textcolor{blue}{//Jointly updating ${\cal M}$ and ${\cal P}_G$}
        \STATE Moderator generates $\nabla {\cal M}_{t}$ using loss ${\cal L}_{server}$.
        \STATE Moderator updates ${\cal M}_{t+1} ={\cal M}_{t} + \nabla {\cal M}_{t}$.
    \ENDFOR
\end{algorithmic}
\end{algorithm}

\begin{figure}[t]
\centerline{
  {\footnotesize (a)}
      \includegraphics[width=0.45\linewidth]{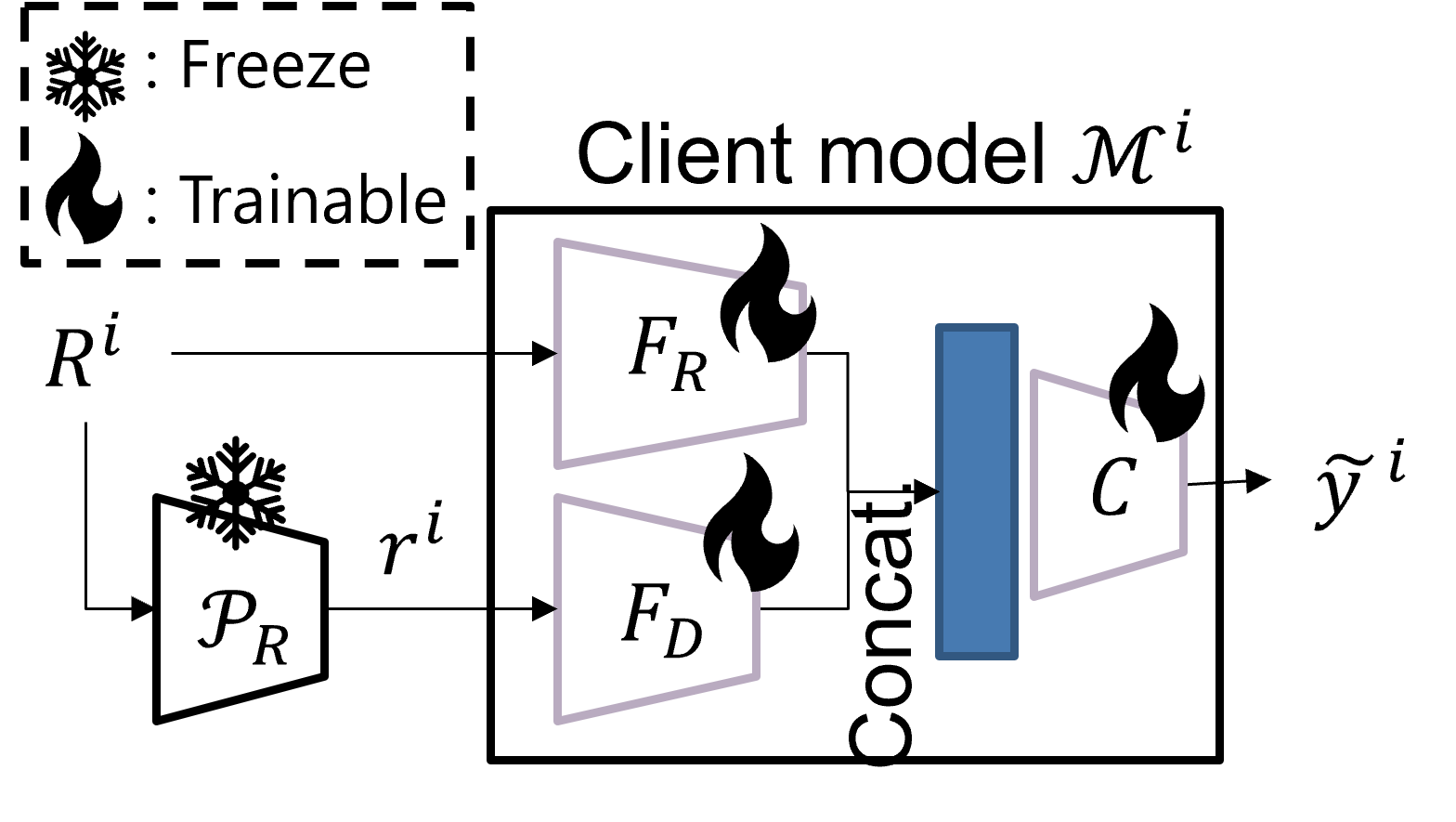}
  {\footnotesize (b)}
  \includegraphics[width=0.45\linewidth]{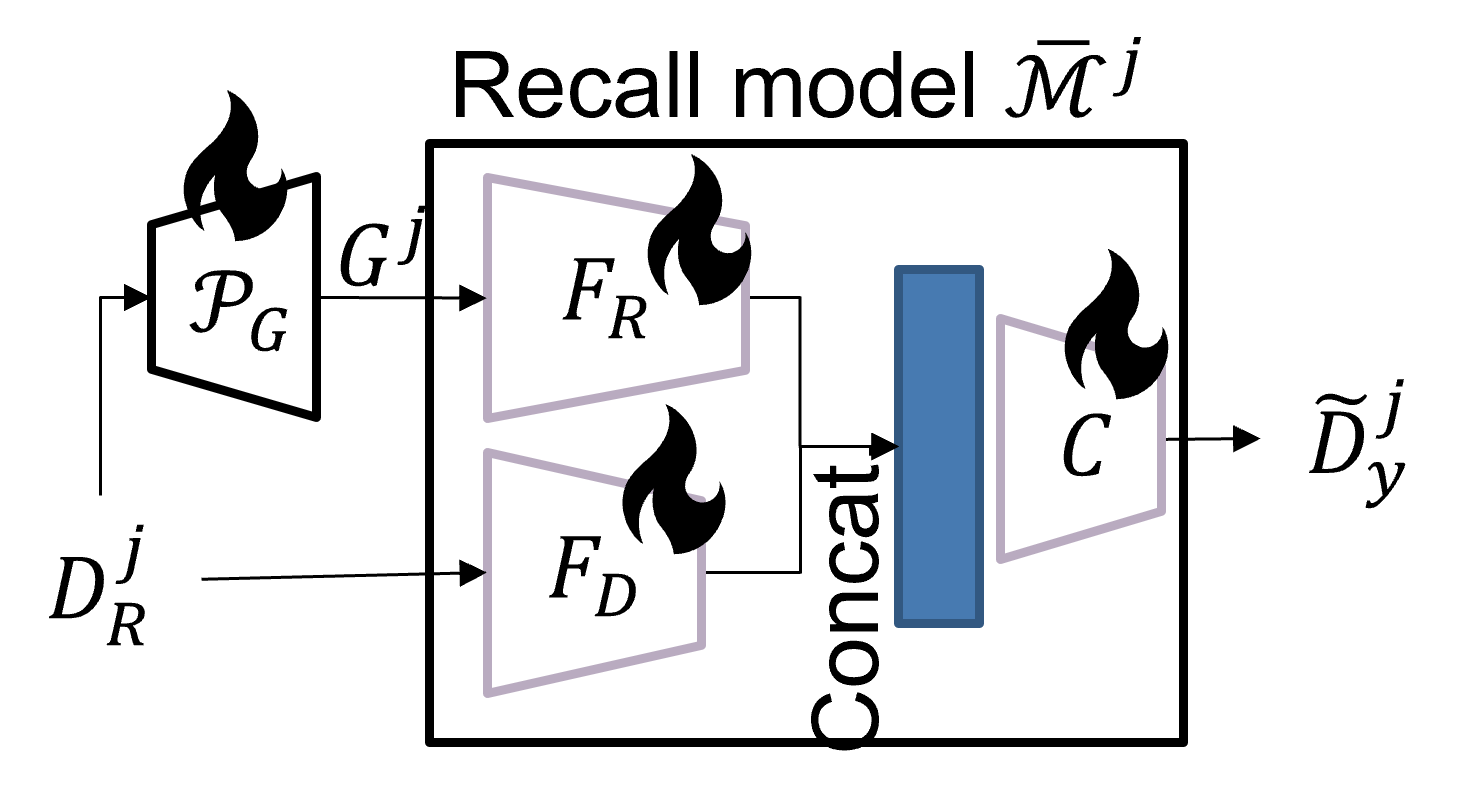}
}
\caption{Training at each client and the moderator:
(a) The client model ${\cal M}^i$ takes raw data $R^i$ and an encoded feature $r^i$ to produce a label $\tilde{y}^i$.
(b) The recall model ${\cal \overline{M}}^j$ takes training guidance $G^j$ and data digest $D^j$ to produce a soft label $\tilde{D}_y^j$.
Feature extractors $F_R$ and $F_D$ extract latent features from the raw data (or guidance) and digests, respectively. $C$ is the classifier producing the final classification.
%
}
\label{fig:model_overview}
\end{figure}

The client and moderator use different types of inputs in FedDig. Clients access raw data, while the server only has access to digests.
Figure~\ref{fig:model_overview} illustrates the training data flow at each client and at the moderator.
(1) For available clients: The client model uses a raw sample $R^i$ and an unmixed feature $r^i$ to produce a hard classification label, where $r^i$ is encoded from the raw sample using feature encoder ${\cal P}_R$. This is shown in Figure~\ref{fig:model_overview}(a).
(2) For absent clients: The moderator generates a recall model that uses the training guidance $G$ and a mixed digest $D_R$ to produce a soft label. Guidance $G$ is derived from digest $D_R$ using the guidance producer ${\cal P}_G$. This is shown in Figure~\ref{fig:model_overview}(b).

The FedDig design maintains consistent architectures across the server model ${\cal M}$, the client model ${\cal M}^i$, and the recall model $\overline{\cal M}^i$.
User can customize the structures of ${\cal P}_R$ and ${\cal P}_G$ based on the specifics and complexity of their FL task.
Both the client and the server models use two feature extractors, $F_R$ and $F_D$, to extract latent features from the raw data (or the guidance) and the digests, respectively. These extracted latent features are then concatenated and fed to the classifier $C$ to produce the final  inference result.

In FedDig, the feature encoder ${\cal P}_R$ is shared among all FL clients, while the guidance producer ${\cal P}_G$ remains a private model of the FL moderator. We initialize ${\cal P}_R$ using the encoder from a pre-trained convolutional autoencoder from~\cite{8095172}, fixing it to ensure that the features $r$ are consistent and that all digests $\cal D$ are static during training. This setup minimizes transmission overhead, as the digests need to be sent only once at the start. The feature encoder ${\cal P}_R$ could be a sophisticated feature extractor, such as the initial layers of a pre-trained model like ResNet18, or any advanced privacy-removing process, such as methods that remove sensitive regions. In our experiments, ${\cal P}_G$ and ${\cal M}$ are initialized randomly.


\subsection{Generating Privacy-Controllable Data Digests}
\label{sec:digest}

We generate privacy-controllable digests with two controllable perturbing methods: feature mixing and differential privacy perturbation.

Starting with encoded features $r$, which already have a low privacy risk due to information loss during feature encoding, we mix multiple feature vectors. We project raw data $R$ into a lower-dimensional feature space using a fixed feature encoder ${\cal P}_R$ and then mix multiple features $\{ r_k \}$ and their corresponding labels $\{ y_k \}$ using a weighted sum, controlled by the hyperparameter Samples per Digest ($SpD$). Each raw sample is accessed only once during maxing, which limits the potential for statistical attacks to reveal information identifying original samples. Each client $i$ thus has $|R^i|/SpD$ mixed features.

Next, we apply the noise-adding mechanism of Differential Privacy~\cite{Fan-DP2018,algo_DP}, $DP(\varepsilon)$, with a hyperparameter $\varepsilon$ to perturb the mixed features in each client. A smaller $\varepsilon$ means the mixed features are harder to distinguish, enhancing privacy.
Each digest $D$ consists of a pair of mixed features $D_R$ with perturbation and soft labels $D_y$, where $D_R = \sum_{k=1}^{SpD}{w_kr_{k}} + DP(\varepsilon)$. Here, $w_k$ represents the mixing weights summing to one. The soft label is calculated by
$D_y=\sum_{k=1}^{SpD}{w_ky_k}$.

We implement $DP(\varepsilon)$ using the Laplace mechanism: $DP(\varepsilon) = Lap (0, \tau^i/(S\varepsilon))$, where $Lap (\mu,\sigma)$ is the Laplace distribution with mean $\mu=0$ and scale $\sigma=\tau^i/(S\varepsilon)$. The scale $\sigma$ combines the global sensitivity $\tau^i/S$ and the parameter $\varepsilon$. Here, $\tau^i$ is set to the largest value among all features held by each client, as  feature values are positive due to the ReLU function. 
Ideally, the global sensitivity should be defined as $\tau^i/|D^i|$, where $|D^i|$ is the client's dataset size. However, in our experiments, we use a constant $S$ (20,000) to standardize the scale magnitude across all clients, while $\varepsilon$ controls the level of perturbation.

\begin{figure}[t]
\centerline{
  \includegraphics[width=1.0\linewidth]{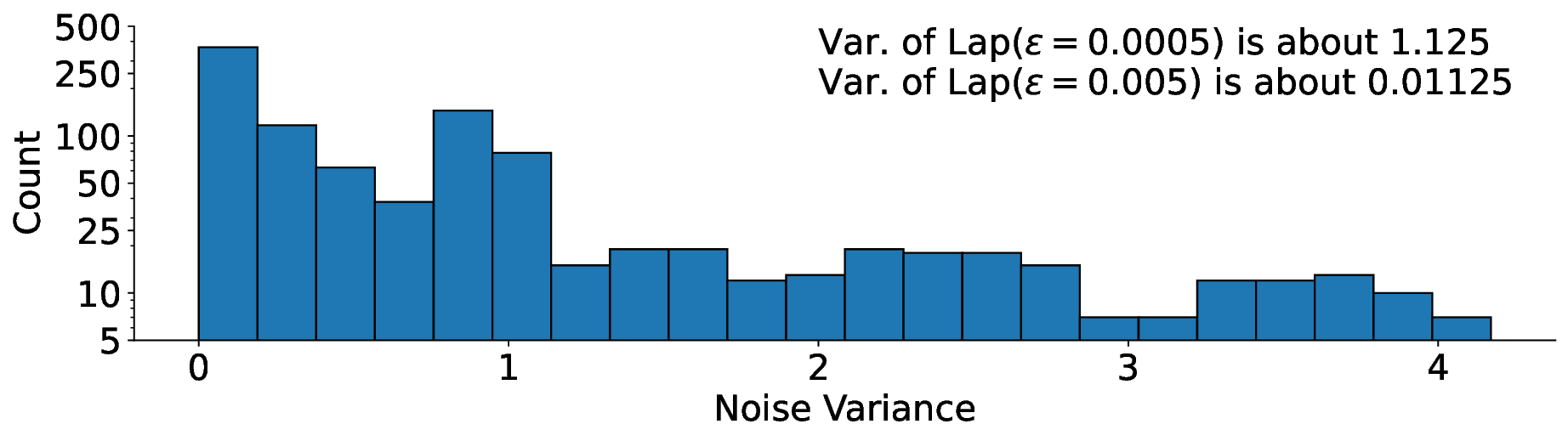}
}
\caption{The noise variances of the mixing method with $SpD=4$.
}
\label{fig:perturb}	
\end{figure}

The digest feature mixup, achieved through the weighted sum of sample mixing. can be understood as {\it perturbing important feature elements with a larger variance}, with the perturbation being derived from the features of other samples. In contrast, the DP noise-adding mechanism uniformly perturbs all feature elements with a fixed variance.
Figure~\ref{fig:perturb} illustrates an example of the distribution of the perturbation ({\em i.e.}, variance of added noises) resulting from the mixup.


\subsection{Data Digest Privacy Assessment}
\label{sec:privacy}

A privacy concern with FedDig is the risk of reverse-engineering the data digests, as they are transmitted beyond the client environment. To address this, we assess the privacy risk of data digests by analyzing both the feature mixing process and the Laplace mechanism of Differential Privacy (DP).

For the {\bf privacy definition}, we focus on preserving {\it intra-class, per-sample privacy}. This means protecting the differentiation between individual samples within a class, rather than enabling inter-class identification (distinguishing samples from different classes). 
This definition aligns with the classification task, as preventing class identification from the digests ensures they cannot be effectively used for model training.

The FedDig algorithm is designed to be $\varepsilon$-differentially private, based on the Laplace mechanism of DP~\cite{DP:2011}. The $\varepsilon$ value serves as a quantifiable measure of data privacy, allowing users to choose the desired level of privacy for their application.
We also examine the feature mixing process, which presents an under-constrained problem where recovery of original features is generally impossible due to information loss. However, it is essential to assess the risk level by calculating the probability $P_{crt}(D)$ that {\em a random guess on a digest $D$ can correctly recover all feature elements before mixing}.  
In summary,  we quantitatively assess the data digest privacy using the pair $(\varepsilon, {\cal P}_{crt}(D))$.



\medskip
\noindent
{\bf Probability in guessing the correct features $r_{k}$: $P_{crt}(D)$}
We discuss a scenario where an adversary monitors the digest transmission and attempts to compromise a recovery model ${\cal P}_R^{-1}$. This model aims to reconstruct the raw data $R_k$ from the digest $D=(D_R,D_y)$, where $D_R = \frac{\sum_{k=1}^{SpD}{w_kr_k}}{SpD}$, $D_y = \frac{\sum_{k=1}^{SpD}{w_ky_k}}{SpD}$, $r_k$ represents the encoded feature indexing by $k$, $w_k$ represents the mixing weight, and $SpD$ is the samples per digest.
We assume that the adversary possesses knowledge of $w_k$ and the digest $D$. Consequently, each unmixed feature can be employed to reconstruct the raw samples, denoted as $R_k={\cal P}_R^{-1}(r_k)$.

Let the digest $D_R$ be an $\ell$-dimension vector, and assume each element $D^e_R$ in $D_R$ is independent.
Subsequently, $P_{crt}(D)$ is the product of the probability of correctly guessing each element, given by $P_{crt}(D)=\prod_{e=0}^{\ell}P_{crt}(D^e_R)$.
In this analysis, we assume that $w_k$ is a constant of $1/SpD$. Consequently, the adversary only needs to guess the correct $r_k$ from $D_R^e = \frac{\sum_{k=1}^{SpD}{r_k^e}}{SpD}$. This signifies that $D_R^e$ is the weighted sum of the $e^{th}$ feature element, formatted as \texttt{float32}.
In a more general context, \texttt{float32} has the potential to be quantized into non-negative integers. Subsequently, the system can be simplified to the summation of these non-negative integers. Given this setup, we compute $P(D_R^e)$ by considering all possible combinations of $r_k^e$ using combinatorics.
Namely,
$P_{crt}(D^e) =
\sum_{V_{D^e}=1}^{I}\frac{1}{I}\left(\frac{1}{\kappa}\right)$,
where $\kappa = C_{V_{D_R^e}}^{SpD+V_{D_R^e}-1} - m$, $C$ denotes the symbol of combination, $V_{D^e}$ is the actual value of $D^e$, which is assumed to be uniform, $I$ is the range of the non-negative integer, and $m$ is the additional counts of the $r_k^e$ permutations. Since the growth rate of $\kappa \geq V_{D_R^e}$ when $SpD\geq3$, we calculate the upper bound of $P_{crt}(D_R^e)$ by
\begin{equation}
\nonumber
P_{crt}(D^e) \leq
\frac{1}{I}\sum_{V_{D_R^e}=1}^{I}\left(\frac{1}{V_{D_R^e}}\right)
\approx \frac{1}{I}\left(ln(I)+\gamma+\frac{1}{2I}\right) \leq \frac{22.8}{2^{32}},
\end{equation}
where $\gamma\approx0.577$ when $I=2^{32}$. As the elements within the digest are independent, $P_{crt}(D_R^e)$ is consistent across all $e$. Consequently, we derive $P_{crt}(D)\leq\left(\frac{22.8}{2^{32}}\right)^\ell$. In our EMNIST experiment, we set $\ell = 256$. This represents $P_{crt}(D)\leq\left(\frac{22.8}{2^{32}}\right)^{256}$, which is extremely small.

While privacy considerations can vary based on specific problems and user preferences, our assessment method provides a foundation for evaluating potential risks. This approach allows users to conduct an initial assessment and better understand the possible privacy implications.

\begin{figure}[t]
\centerline{
  \includegraphics[width=0.5\textwidth]{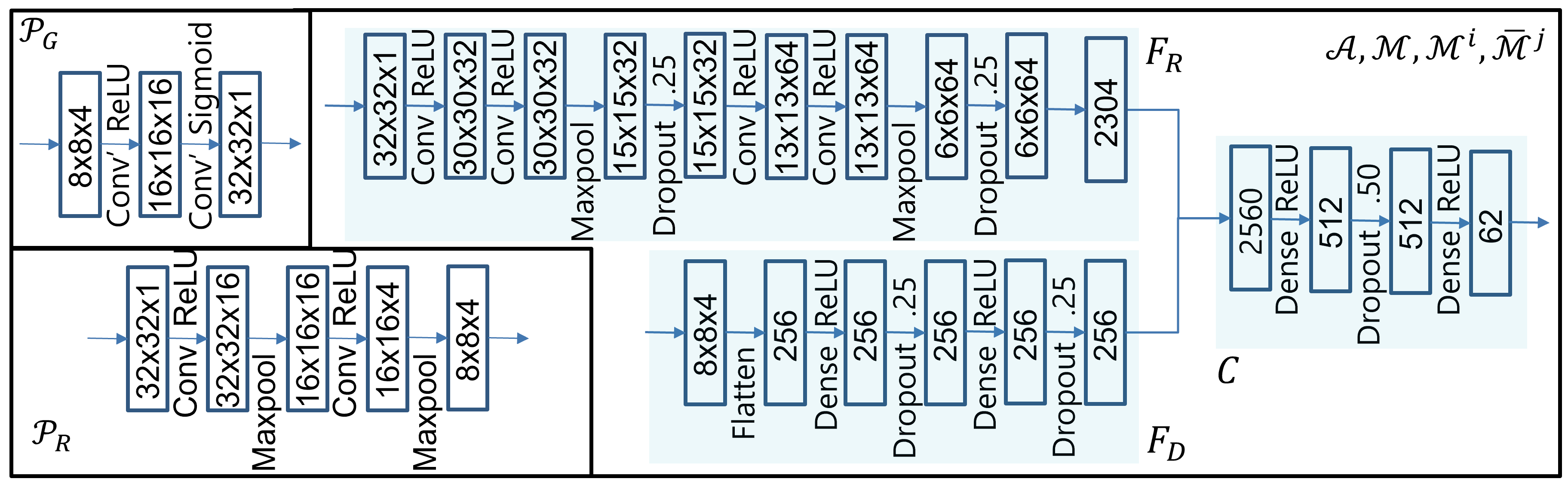}
}
\caption{
Model structures used in the EMNIST experiment.
}
\label{fig:architectures}	
\end{figure}

\begin{figure*}[t]
\centerline{
  {\scriptsize (a)}
  \includegraphics[width=0.175\linewidth]{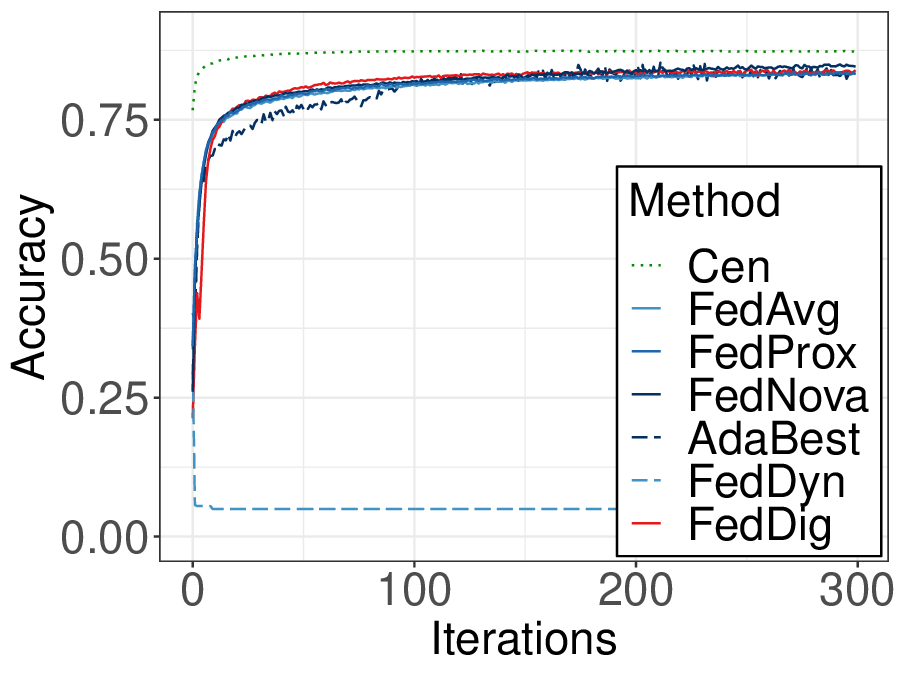}
  {\scriptsize (b)}
  \includegraphics[width=0.175\linewidth]{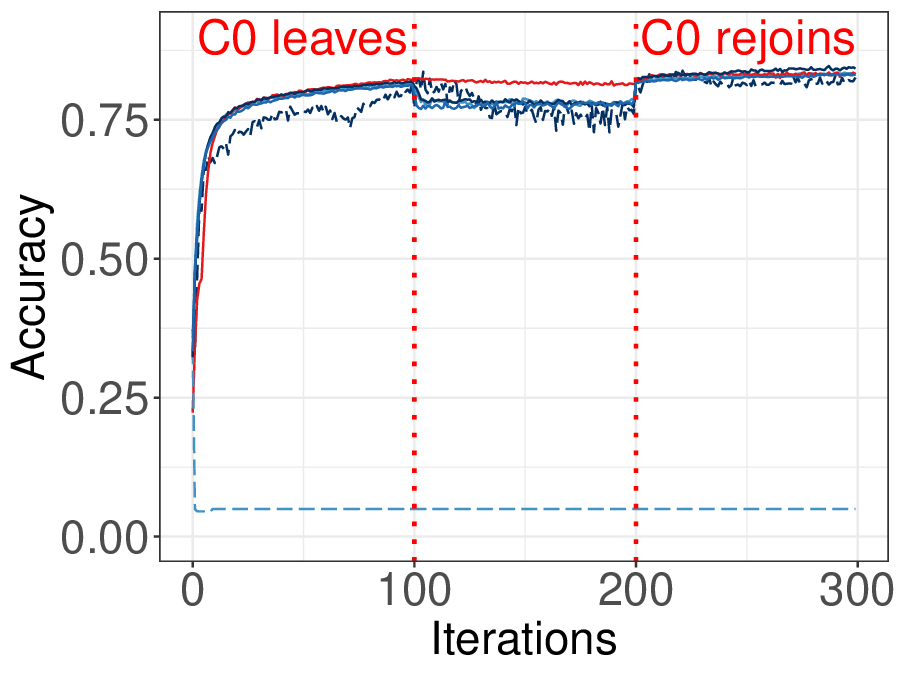}
  {\scriptsize (c)}
  \includegraphics[width=0.175\linewidth]{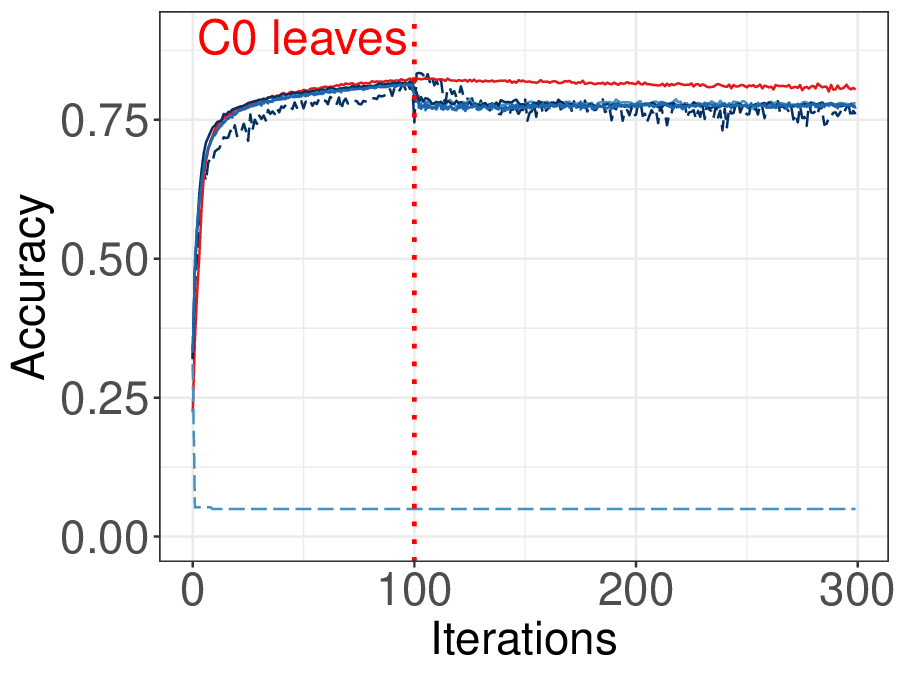}
  {\scriptsize (d)}
  \includegraphics[width=0.175\linewidth]{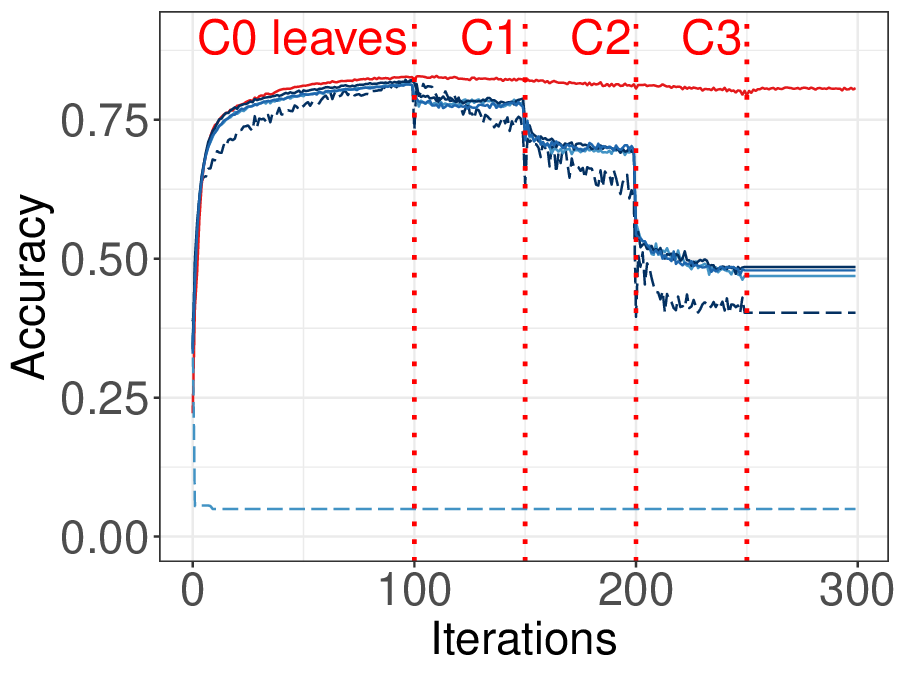}
  {\scriptsize (e)}
  \includegraphics[width=0.175\linewidth]{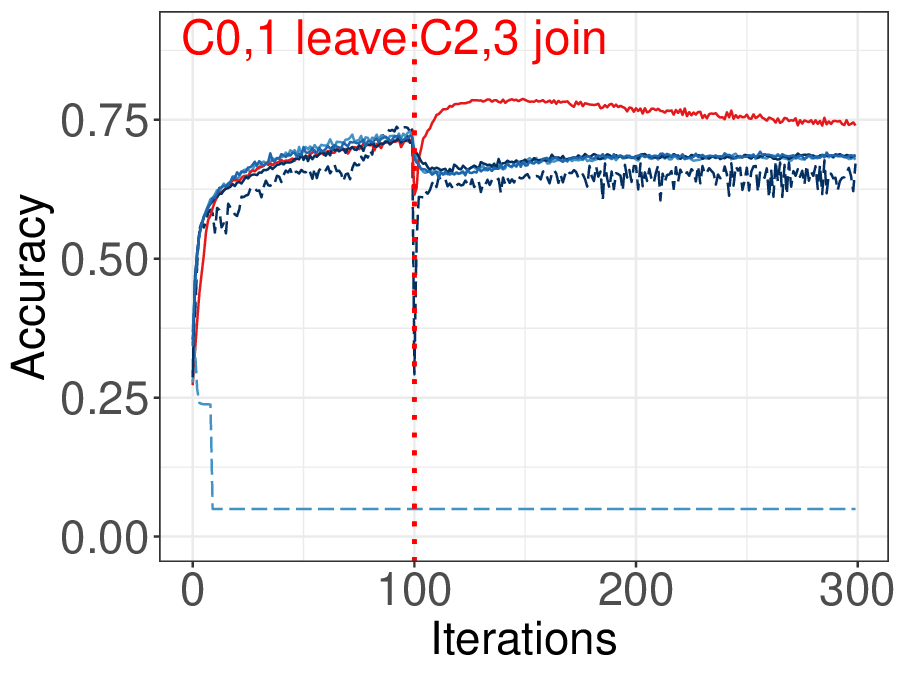}
}
\caption{Compare FedDig with baselines for the four client absence scenarios in the EMNIST experiments: (a) No client leaves. (b) The largest client leaves temporarily. (c) The largest client leaves permanently. (d) All clients leave sequentially. 
(e) Two groups of clients join the training at different times.}
\label{fig:Scenarios}	
\end{figure*}
\begin{figure}[t]
\centerline{
  \begin{subfigure}[b]{0.5\linewidth}
         \centerline{
         \includegraphics[width=\linewidth]{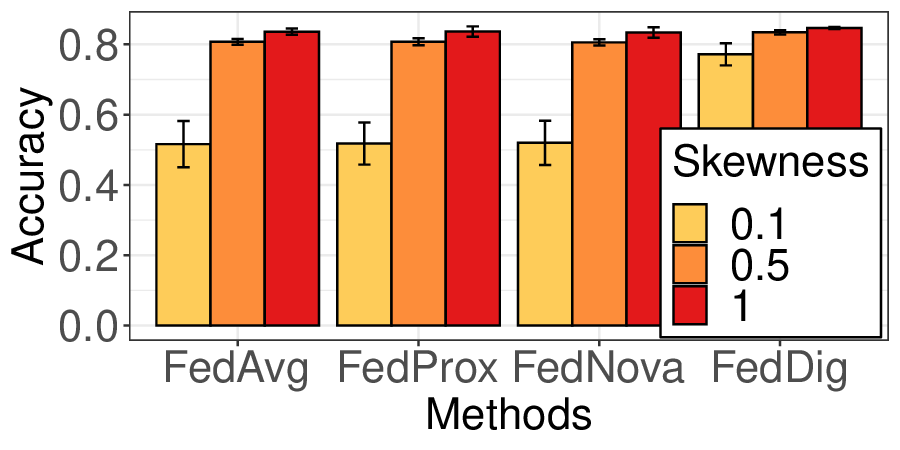}
         }
         \caption{Skewness}
  \end{subfigure}
  \begin{subfigure}[b]{0.5\linewidth}
         \centerline{
         \includegraphics[width=\linewidth]{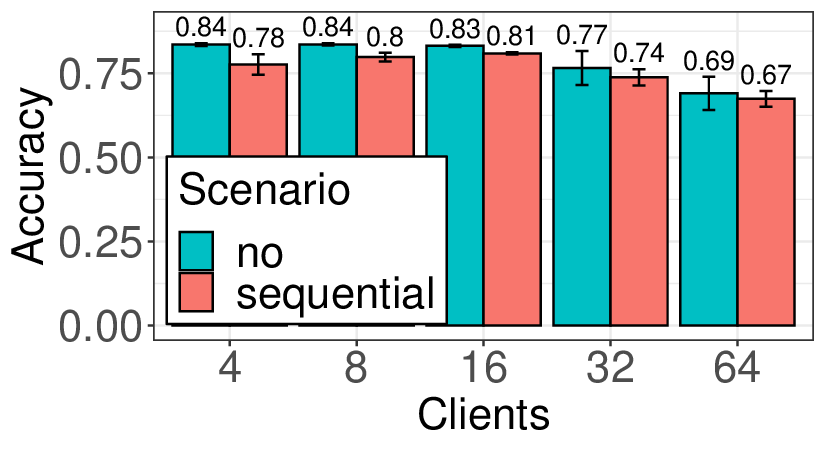}
         }
         \caption{\# of Clients}
  \end{subfigure}
}
\caption{
(a) The mean and standard deviation of test accuracy from training on various data distributions for sequential leaving scenario on EMNIST.
(b) The mean and standard deviation of test accuracy under different number of clients.
}
\label{fig:distrubance}	
\end{figure}
\section{Experimental Results}
\label{sec:exp}

We evaluate FedDig using standard image classification datasets, including EMNIST ByClass~\cite{cohen2017emnist}, Fashion-MNIST~\cite{fmnist}, CIFAR-10, and CIFAR-100~\cite{cifar}, to simulate and investigate the scenarios of FL client absence during the training  deep neural network models. We compare FedDig with five popular FL algorithms: FedAvg~\cite{FedAVg:AISTATS2017}, FedProx~\cite{FedProx:MLSys2020}, FedNova~\cite{FedNova:NeurIPS2020}, FedDyn~\cite{FedDyn:ICLR2021}, and AdaBest~\cite{AdaBest}, as these are widely used in FL for handling different client data contributions.

\medskip
\noindent
{\bf Experimental Settings:}
All experiments are conducted on a Linux server with six Intel\textsuperscript{\scriptsize\textregistered} Xeon\textsuperscript{\scriptsize\textregistered} CPU E5-2690 v4 2.60GHz, 115GB RAM, and a V100 GPU. Implementation are done using PyTorch 1.10.1+cu102.
The FL clients are trained sequentially due to hardware limitations. We ensure the convergence of the server model by training all client models for 300 iterations without early stopping. In each iteration, client models are trained on their respective raw data for 1 epoch. The batch size is set to 256 for E/FMNIST and 32 for CIFAR-10/100 datasets. We use a learning rate of 0.001 with Stochastic Gradient Descent (SGD) and 0.9 momentum. The $SpD$ defaults to 4. Generally, we select the value empirically to balance the trade-off between accuracy and privacy level. The size of training samples, $S$, is set to 20,000 for E/FMNIST and 10,000 for CIFAR-10/100. The number of clients and the split of the training/validation/test set match those used in the pilot experiment described in the Preliminary Section. Training samples are distributed to clients according to a Dirichlet distribution with parameter $\varepsilon = 0.005$. The data skewness parameter $\mu$ is set to 0.1 unless otherwise specified. The evaluation metric used is the average test accuracy. Figure~\ref{fig:architectures} shows the network architectures.




\subsection{Impact of Client Leaving and Data Distribution}

We compare several FL algorithms under conditions of high data skewness, imbalanced, and non-IID distributions to evaluate the effectiveness of FedDig. Specifically, clients 0, 1, 2, and 3 possess 34.1\%, 23.6\%, 23.8\%, and 18.5\% of the unbalanced training samples, respectively.

We test four client absence scenarios using the EMNIST dataset: (1) The largest client (holding the most samples) leaves temporarily. (2) The largest client leaves permanently. (3) All clients leave sequentially.  (4) Two groups of clients join the training at different times. For reference, the average test accuracy under centralized training is 87$_{\pm0.02}$, which serves as the upper bound. The value following the $\pm$ symbol indicates the standard deviation across five experiments with different random seeds.

The {\em test accuracy} results, illustrated in Figure~\ref{fig:Scenarios}, demonstrate the impact of client absence across different scenarios.
In scenarios 1 through 3 (Figures~\ref{fig:Scenarios}(b,c,d)), baseline algorithms experience catastrophic knowledge forgetting due to the non-IID nature of the training samples once clients are absent.
In contrast, FedDig maintains stable test accuracy even when all clients are absent, showcasing the effectiveness of using data digests and training guidance. Notably, FedDyn fails to converge in these scenarios, with test accuracy plummeting to 4.97\% within the initial iterations.

Figure~\ref{fig:Scenarios}(e) presents an interesting case, where the clients are divided into two groups, with each group joining the FL training separately. During the first 100 iterations, all algorithms show poor performance (73\% test accuracy) due to insufficient training samples for certain classes. However, once the second group joins, FedDig's accuracy significantly improves, benefiting from the digests and guidance. This indicates that the digests retain valuable information about the first group's data. In comparison, the baseline algorithms struggle with data absence.

Table~\ref{tab:all_scenario} summarizes the overall test accuracy for each scenario, where FedDig generally outperforms the baseline algorithms.


\begin{table}[t]
\caption{Performance comparison on challenging scenarios. $^*$FedDyn can not converge in all scenarios.}
\label{tab:all_scenario}
\centering
\begin{tabular}{crrrrr}
\hline
\multicolumn{1}{l}{} & \multicolumn{1}{c}{No} & \multicolumn{1}{c}{Temp.}  & \multicolumn{1}{c}{Forever} & \multicolumn{1}{c}{Seqent.} & \multicolumn{1}{c}{Group} \\ \hline
FedAvg   & 84.6      & 83.0       & 77.5       & 51.6       & 68.3             \\ \hline
FedProx  & 84.6      & 82.7       & 77.5       & 51.8       & 68.6             \\ \hline
FedNova  & \bf{85.0} & \bf{83.8}  & 77.7       & 53.0       & 68.4             \\ \hline
FedDyn   & 5.0$^*$   & 5.0$^*$    & 5.0$^*$    & 5.0$^*$    & 5.0$^*$               \\ \hline
AdaBest  & 83.5      & 82.7       & 76.6       & 47.0       & 68.1             \\ \hline
FedDig   & 83.5      & 83.2       & \bf{81.0}  & \bf{77.6}  & \bf{75.3}   \\ \hline
\end{tabular}
\end{table}

\subsection{Impacts on FedDig Hyperparameters} 
\label{sec:disturbance}

We conduct experiments to investigate the impact of data distribution and the FedDig hyperparameters. From this section onward, all experiments are conducted under the sequential leaving scenario, with $SpD$ set to four unless otherwise specified. We compare FedDig against FedAvg, FedProx, and FedNova, as they perform better in the sequential leaving scenario. The evaluation metric is the average test accuracy after all clients have left (from iterations 251 to 259), calculated across five repeated experiments with different random seeds on the EMNIST dataset. The error bars represent the standard deviation of these five experiments.

\medskip
\noindent
{\bf Non-IID data skewness level:}
Figure~\ref{fig:distrubance}(a) presents the impact of varying the Dirichlet distribution parameter $\mu$ at levels $0.1$, $0.5$, and $1.0$. We observe that FedDig maintains stable accuracy across different skewness levels. In contrast, the other baseline algorithms experience varying degrees of accuracy degradation as the data skewness increases. Notably, FedDig significantly outperforms all baseline algorithms when $\mu = 0.1$, demonstrating that {\it FedDig effectively manages various data distribution levels.}

\medskip
\noindent
{\bf Number of clients:}
To assess the impact of the number of clients, we randomly select 25\% of clients to sequentially leave the training at iterations 100, 150, 200, and 250. Figure~\ref{fig:distrubance}(b) presents the results for scenarios with 4 to 64 clients, comparing situations where no clients leave (``no'') versus all clients sequentially leaving (``sequential''). 
We observe that accuracy declines when the number of clients exceeds 32. 
However, the accuracy difference between the no-leaving and sequential-leaving scenarios remains relatively constant at around 3\%, indicating FedDig's stability. This suggests that FedDig's performance is more influenced by the amount of missing data rather than the number of clients in cross-silo FL. Therefore, {\it FedDig can accommodate varying numbers of clients without significant limitations and has the potential to support large-scale scenarios,} including those with more than 64 clients.



\medskip
\noindent
{\bf Backbone FL algorithms:}
We investigate the effects of integrating FedDig with other FL algorithms, specifically FedProx and FedNova.
As in previous experiments, we evaluate the test accuracy from five runs with different seeds.
When using FedAvg as the base, FedDig improves the test accuracy significantly, from $51.6\%_{\pm6.6\%}$ to $77.6\%_{\pm3.0\%}$.
Similarly, applying FedDig to FedProx and FedNova results in notable accuracy increases, from $51.8\%_{\pm6.0\%}$ to $77.2\%_{\pm2.9\%}$ for FedProx and from $52.0\%_{\pm6.3\%}$ to $80.0\%_{\pm0.9\%}$ for FedNova.
These results demonstrate that {\it FedDig is versatile and adaptable to various FL algorithms, significantly enhancing their performance}.

\begin{figure*}[t]
\centerline{
    \begin{subfigure}[b]{0.3\linewidth}
         \centerline{
         \includegraphics[width=\linewidth]{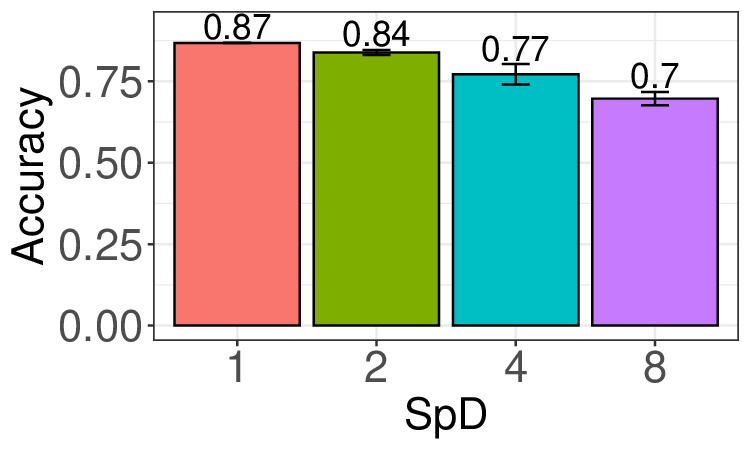}
         }
         \caption{}
  \end{subfigure}
  \begin{subfigure}[b]{0.24\linewidth}
         \centerline{
         \includegraphics[width=\linewidth]{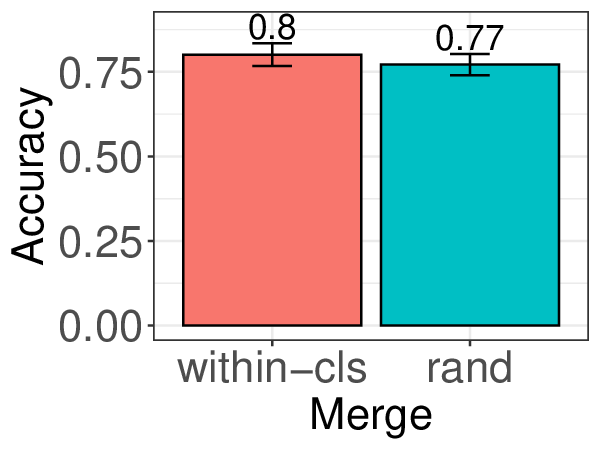}
         }
         \caption{}
  \end{subfigure}
  \begin{subfigure}[b]{0.36\linewidth}
         \centerline{
         \includegraphics[width=\linewidth]{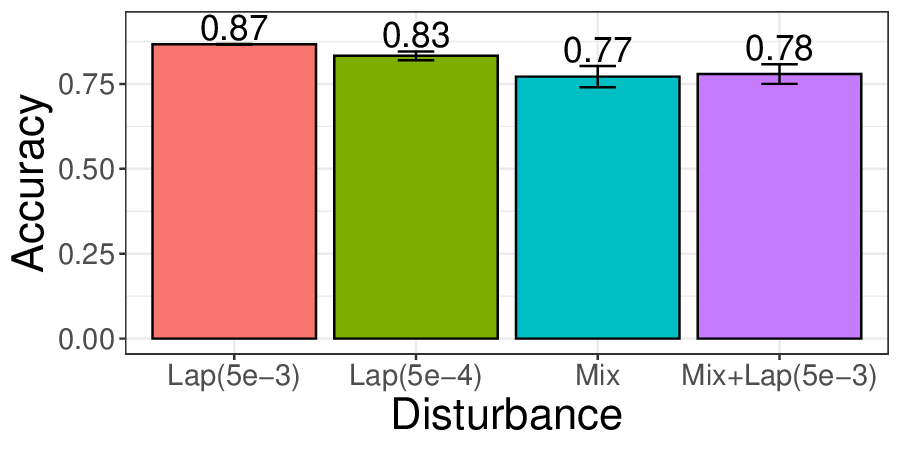}
         }
         \caption{}
  \end{subfigure}
}
\caption{
(a-c) The mean/std. of test accuracy under different sample feature perturbation settings: (a) the $SpD$ value, (b) mixing strategy, and (c) perturbing methods.
}
\label{fig:hyperparameters}
\end{figure*}

\medskip
\noindent
{\bf Samples per Digest.} In Figure~\ref{fig:hyperparameters}(a), the influence of $SpD$ is depicted. A larger $SpD$ results in a reduced test accuracy, which is expected as the perturbation level increases when more samples are mixed. The $SpD$ parameter governs the privacy level of the digest. In the simplistic scenario of $SpD = 1$, the digest is nearly unprotected, and raw data recovery is feasible, making it applicable in cases where privacy is not a concern during client absence. {\it An optimal trade-off between the desired privacy level and model performance is achieved when $SpD = 4$}.

\begin{figure}[t]
    \centerline{
     \includegraphics[width=1.0\linewidth]{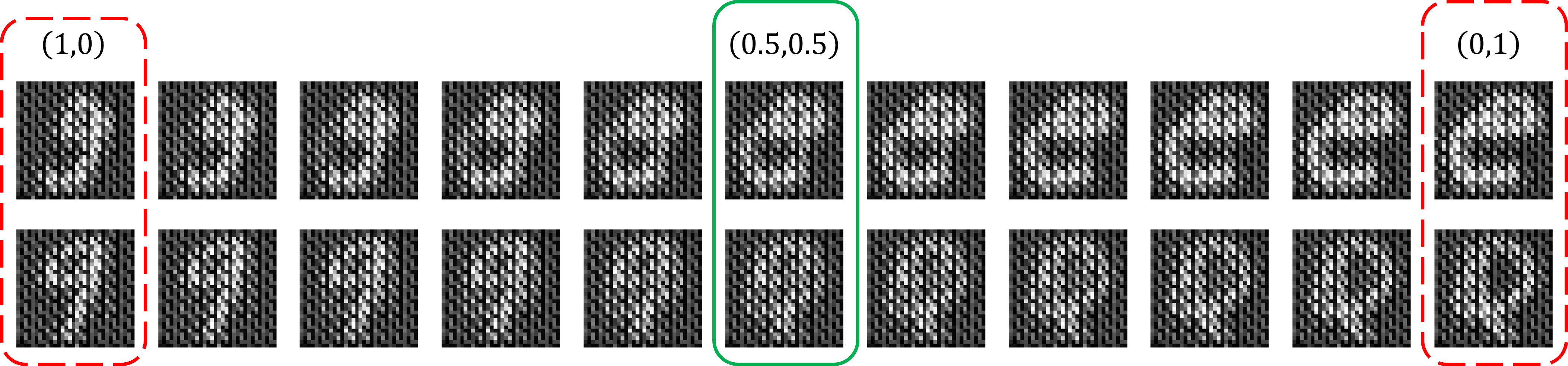}
    }
\caption{The guidance produced with different weights on $SpD=2$.}
\label{fig:weight}	
\end{figure}

\medskip
\noindent
{\bf Mixing weights.} We next discuss the effect of weights in feature mixing. Experiments are conducted with $SpD = 2$ for better visualization. Figure~\ref{fig:weight} shows the guidance produced with weight pairs ranging from $(1,0)$ to $(0,1)$. For the weight pairs of $(1,0)$ or $(0,1)$, the guidance is equivalent to the result of $SpD = 1$. The digest is only protected by the information loss caused by the dimension reduction and the DP scheme. When the weights become balanced $(0.5, 0.5)$, the guidance is most distinct from either side. Such {\it balance weights yield the best digest protection}.

\begin{figure}[t]
      \begin{subfigure}[b]{0.22\linewidth}
         \centerline{
         \includegraphics[width=\linewidth]{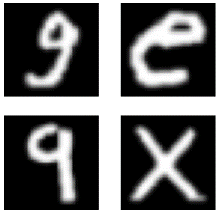}
         }
         \caption{raw data}
         \label{fig:raw_emnist}
     \end{subfigure}
     \hfill
     \begin{subfigure}[b]{0.22\linewidth}
         \centerline{
         \includegraphics[width=\linewidth]{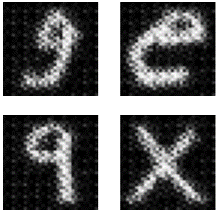}
         }
         \caption{$\varepsilon = 5e^{-3}$}
         \label{fig:e100}
     \end{subfigure}
     \hfill
     \begin{subfigure}[b]{0.22\linewidth}
         \centerline{
         \includegraphics[width=\linewidth]{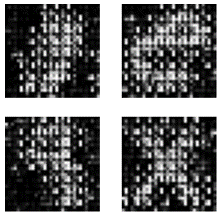}
         }
         \caption{$\varepsilon = 5e^{-4}$}
         \label{fig:e10}
     \end{subfigure}
     \hfill
     \begin{subfigure}[b]{0.22\linewidth}
         \centerline{
         \includegraphics[width=\linewidth]{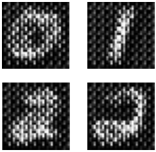}
         }
         \caption{within-cls}
         \label{fig:dist_byclass}
     \end{subfigure}
     \hfill
\caption{Comparison of the raw EMNIST data and the training guidance $G$ produced by the tested approaches.}
\label{fig:noise}
\end{figure}

\medskip
\noindent
 {\bf Mixing strategy.}
We investigate two feature mixing strategies: (1) mixing random features regardless of classes (rand), and (2) mixing random features within each class (within-cls).
Figure~\ref{fig:hyperparameters}(b) shows that no significant difference between the two strategies is found. However, randomly mixing could potentially achieve a higher protection level. Figure~\ref{fig:dist_byclass} shows the guidance produced by within-cls, where the characters are visually recognizable although the original raw data may not be recovered. It is again a trade-off: {\it the security protection of mixing across classes comes with a slight performance drop}.

\begin{figure*}[t]
\centerline{
  \includegraphics[width=.95\linewidth]{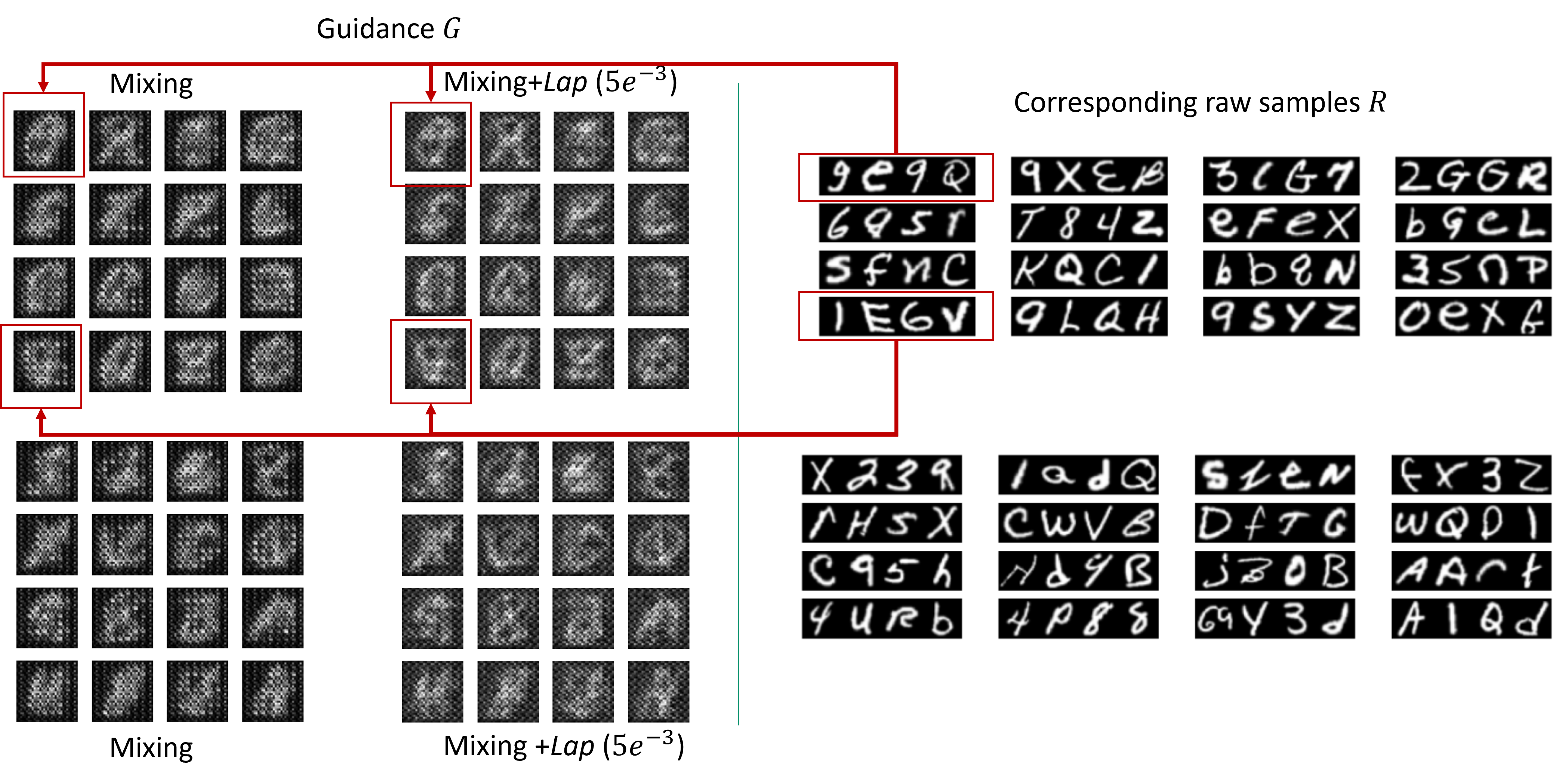}
}
\caption{Examples of the guidance $G$ and the corresponding raw samples $R$ in the EMNIST experiment.}
\label{fig:exp_G}
\end{figure*}

\medskip
\noindent
{\bf Perturbing methods.}
We investigate the impact caused by data mixing and perturbation.
Figure~\ref{fig:hyperparameters}(c) shows the DP-only perturbation achieves better test accuracy. However, high model performance comes with an inferior security level when $\varepsilon$ is small. How to select an appropriate $\varepsilon$ to protect privacy is still an open question for DP-based schemes. The rightmost bar in Figure~\ref{fig:hyperparameters}(c) shows the accuracy of the combination of feature mixing and DP. The accuracy does not show a significant drop. Figure~\ref{fig:noise} shows the visual appearance of some generated guidance samples. Observe that the raw data is visually recoverable by the guidance producer ${\cal P}_G$ in Figure~\ref{fig:e100}, which is unacceptable for privacy protection. Selecting a small $\varepsilon$ could mitigate the privacy issue (Figure~\ref{fig:e10}). However, such noise could be reduced by noise removal. Selecting the parameter for the DP-based methods is difficult because it depends on the combination of the data and the target application. Introducing feature mixing can trade accuracy with better protection over DP-only schemes.
Figure~\ref{fig:exp_G} shows the guidance on the $SpD=4$ experiments. Observe that the characters are barely recognizable from the guidance. {\it FedDig can support a wide range of mixing and perturbation methods}.

\subsection{Experiments on Heterogeneous Datasets}
\label{sec:cifar}

\begin{figure}[t]
\centerline{
      \begin{subfigure}[b]{0.28\linewidth}
         \centerline{
         \includegraphics[width=1.1\linewidth]{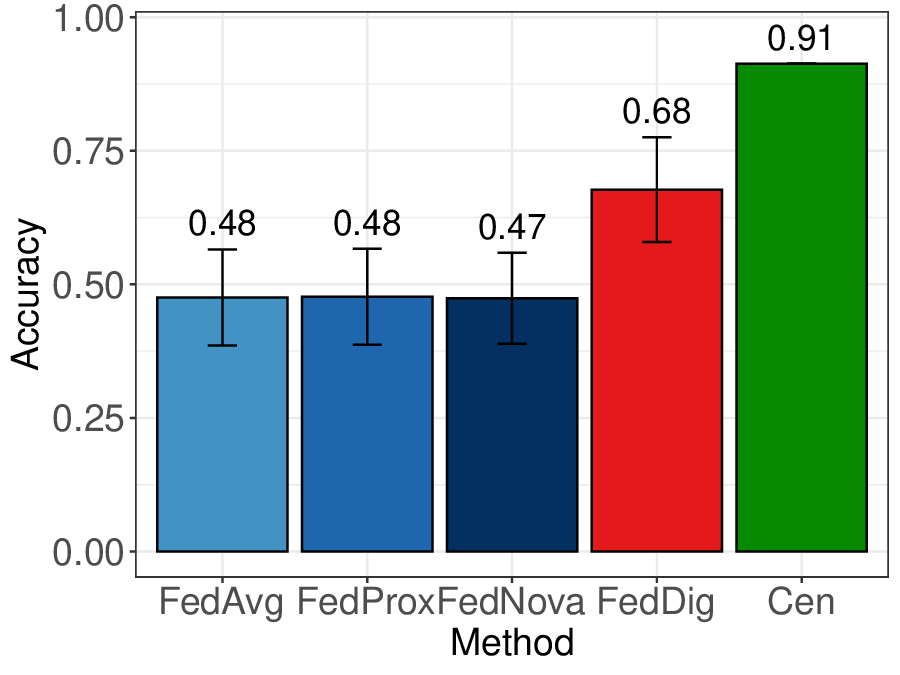}
         }
         \caption{FMNIST}
         \label{fig:bar_fmnist}
     \end{subfigure}
     \hfill
     \begin{subfigure}[b]{0.28\linewidth}
         \centerline{
         \includegraphics[width=1.1\linewidth]{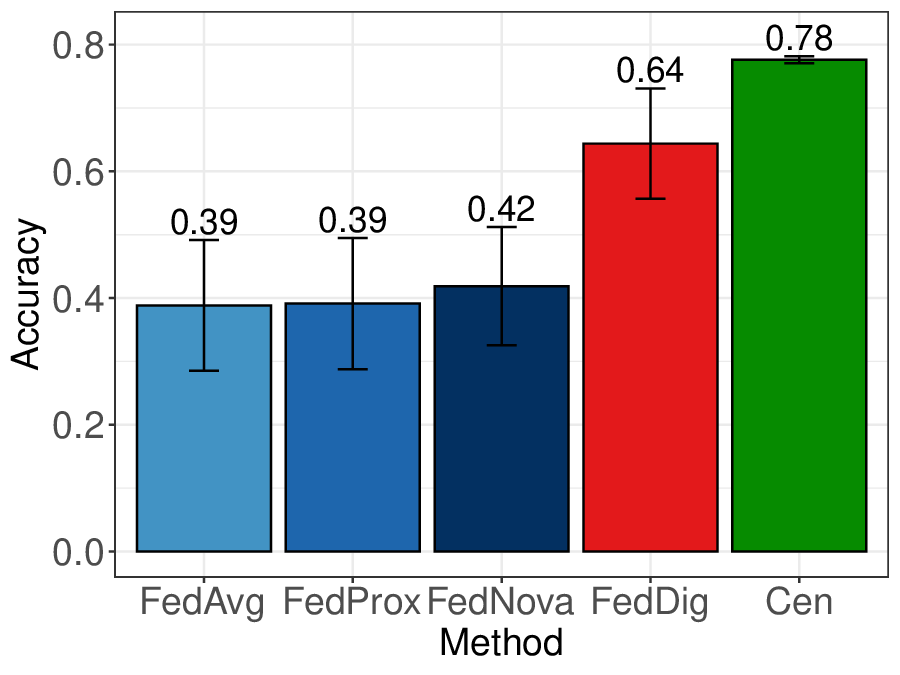}
         }
         \caption{CIFAR-10}
         \label{fig:bar_cifar10}
     \end{subfigure}
     \hfill
      \begin{subfigure}[b]{0.28\linewidth}
         \centerline{
         \includegraphics[width=1.1\linewidth]{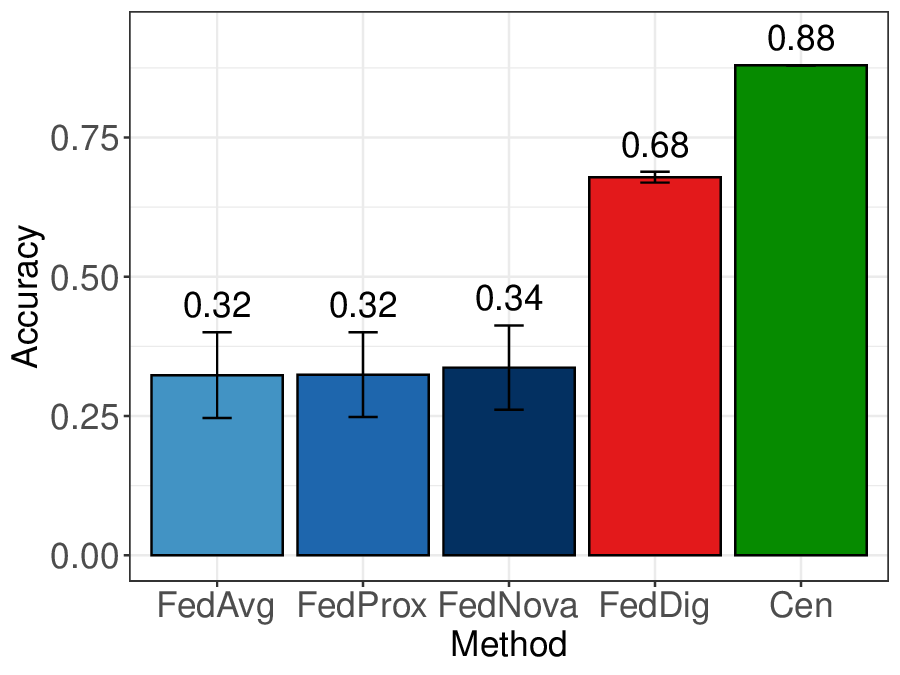}
         }
         \caption{CIFAR-100}
         \label{fig:bar_cifar100}
     \end{subfigure}
    }
\caption{Performance comparison on multiple datasets.}
\label{fig:cifar10_and_cifar100}
\end{figure}

FedDig's scalability is assessed using FMNIST, CIFAR-10, and CIFAR-100 datasets. 
Figure~\ref{fig:cifar10_and_cifar100} shows the mean and standard deviations of test accuracy from five repeated sequential leaving experiments.
For FMNIST and CIFAR-10, the client and server models match those used in the EMNIST experiments to avoid biases due to network architecture differences. The kernel size in the first convolutional layer of $F_R$ is adjusted to 3$\times$3$\times$3 for color images.
As demonstrated in Figure~\ref{fig:bar_fmnist} and Figure~\ref{fig:bar_cifar10}, FedDig significantly outperforms the baseline methods.

For CIFAR-100, we explore the benefits of using highly refined features, which typically prevent raw data recovery due to information loss and reduce transmission overhead when sending digests to the server.
We use the layers before the avgpool layer of 
MobileNetV2~\cite{MobileNetV2} as ${\cal P}_R$, and ${\cal P}_G$ is implemented as a three-layer deconvolutional neural network. For comparison, we directly train MobileNetV2 for FedAvg, FedProx, and FedNova experiments.
Figure~\ref{fig:bar_cifar100} shows that FedDig maintains a significant performance advantage over the baselines. This indicates that {\it FedDig is versatile enough to handle different datasets and that combining highly refined features with raw images can potentially enhance accuracy.}



\section{Guidance visualization}
\begin{figure}[t]
    \centerline{
     \includegraphics[width=1.0\linewidth]{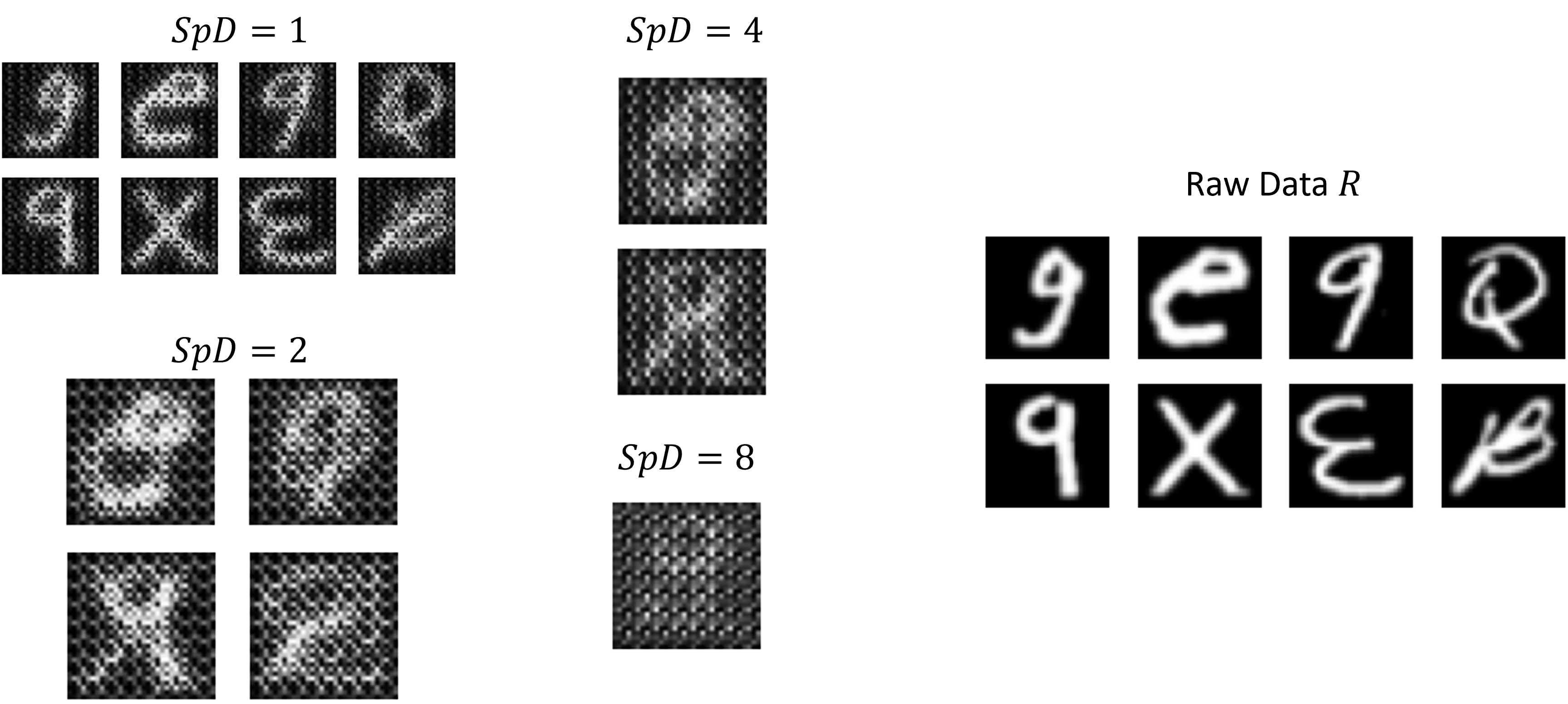}
    }
\caption{The guidance examples with varying $SpD$ values in the EMNIST experiment.}
\label{fig:dif-spd}	
\end{figure}

Figure~\ref{fig:exp_G} shows the training guidance in the EMNIST experiment with $SpD=4$. We analyze the guidance generated from the EMNIST experiment to investigate the impact of $SpD$. Figure~\ref{fig:dif-spd} illustrates the produced guidance for $SpD=1,2,4,$ and $8$. 
Notably, with $SpD=1$, the content is minimally protected as the digest primarily relies on the safeguard provided by the DP Laplace mechanism. As discussed in the paper, choosing an appropriate DP hyperparameter $\varepsilon$ could potentially enhance this protection. However, determining a universally suitable value for different tasks remains an open question.  With an increase in $SpD$, the guidance becomes visually unrecognizable to humans. In summary, the guidance may be deemed acceptable when $SpD=4$.

\begin{figure}[t]
     \centerline{
      \begin{subfigure}[b]{\linewidth}
         \centerline{
         \includegraphics[width=\linewidth]{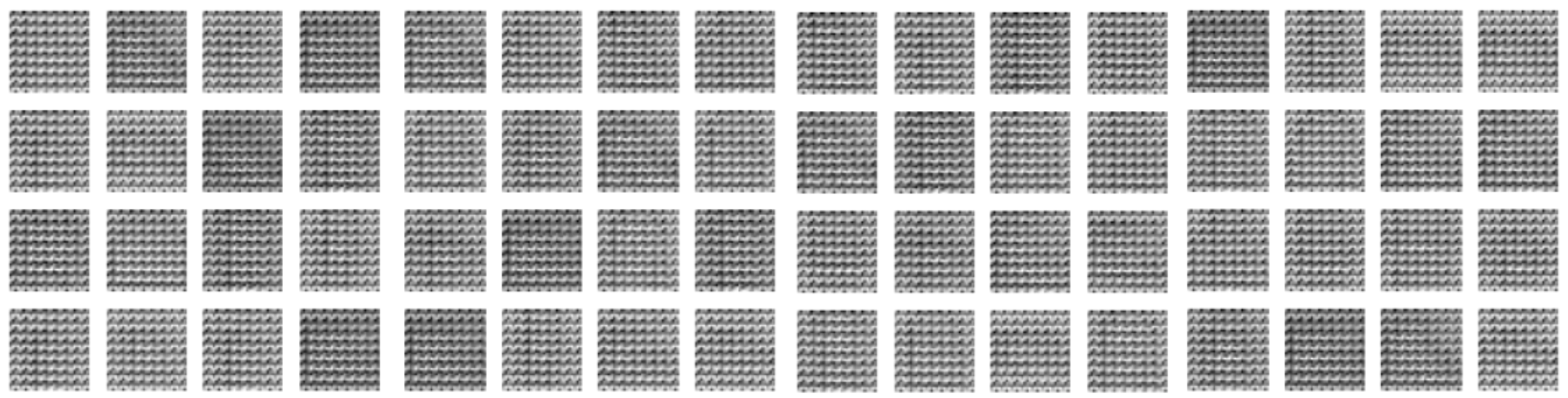}
         }
         \caption{FMNIST}
     \end{subfigure}
     }
     \hfill
     \centerline{
      \begin{subfigure}[b]{\linewidth}
         \centerline{
         \includegraphics[width=\linewidth]{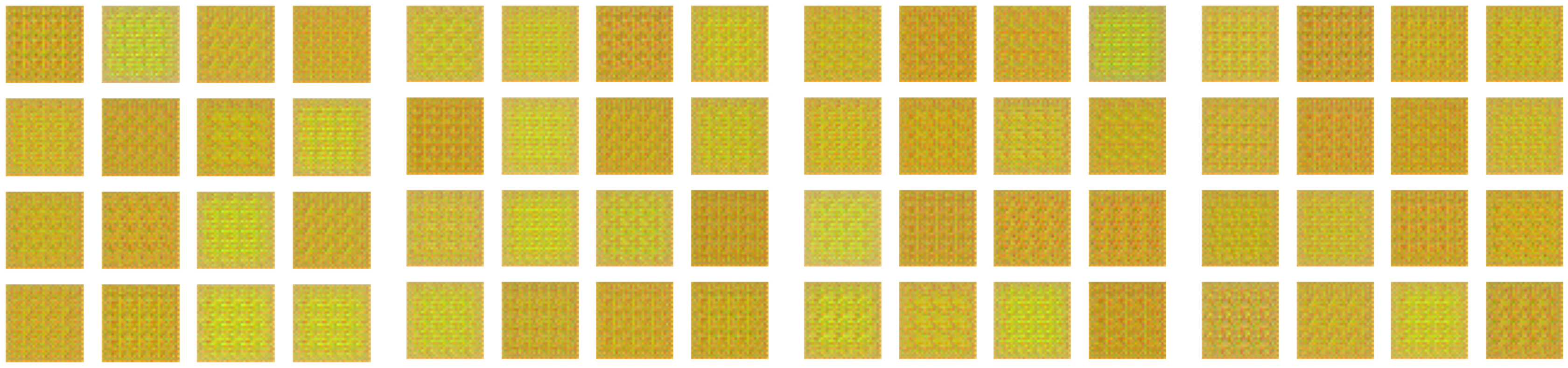}
         }
         \caption{CIFAR-10}
     \end{subfigure}
    }
\caption{Examples of the guidance produced by ${\cal P}_G$ on (a) FMNIST and (b) CIFAR-10.
}
\label{fig:guidance_vis}
\end{figure}

Figure~\ref{fig:guidance_vis} illustrates the $4\textrm{-}SpD$ guidance from the FMNIST and CIFAR-10 experiments, respectively. The guidance for the EMNIST experiment can be found in Figure~\ref{fig:dif-spd} of the paper. In general, the guidance $G$ is visually unrecognizable to humans.. Overall, we can observe that {\it the characters are barely recognizable from the training guidance}.

\section{Direct Reverse Engineering on the Digest.}
We investigate to what extent the digests can be protected under reverse engineering.
Recall that the moderator must transmit ${\cal P}_R$ to all clients to generate the data digest. We assume that an attacker intercepts this transmission and compromises ${\cal P}_R$ to obtain its {\em pseudo-inverse} ${\cal P}_R^{-1}$ with the intent of recovering the raw data from the digest directly.

\begin{figure}[t]
     \centerline{
     \includegraphics[width=\linewidth]{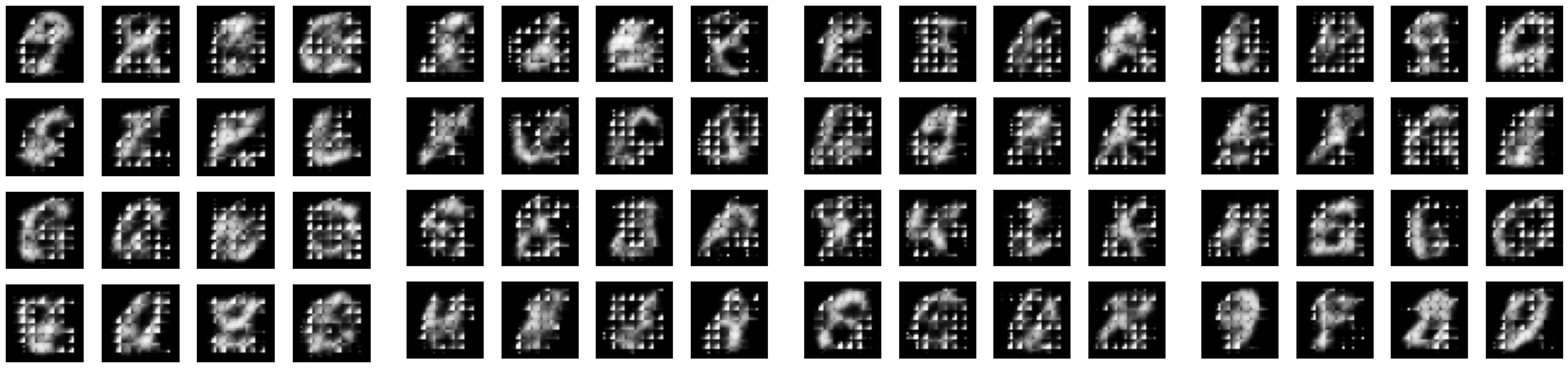}
     }
\caption{Recovered Examples with the adversarial pseudo-inverse ${\cal P}_R^{-1}(D_R)$ on EMNIST.
}
\label{fig:recover}	
\end{figure}

We replicate the reverse engineering experiment on the EMNIST dataset by training {\em autoencoders} with the same structures as ${\cal P}_R$ and ${\cal P}_G$. Similarly, for the CIFAR-10 dataset, we employ the MobileNetV2-based ${\cal P}_R$ and the corresponding ${\cal P}_G$. Subsequently, we utilize the encoder part as ${\cal P}_R$ and the decoder part as ${\cal P}_R^{-1}$. The recovered results by ${\cal P}_R^{-1}$ are then visualized.
Figure~\ref{fig:recover} shows the recovered result of the EMNIST experiment. Although the recovered results ${\cal P}_R^{-1}(D_R)$ reveal some patterns, individual raw data are far from identifiable from these pattern-like samples, thanks to the {\em unrecoverable} feature mixing. Overall, the recovered results are also visually unrecognizable to humans in all our tested datasets.

\begin{figure}[t]
    \centerline{
     \includegraphics[width=1.0\linewidth]{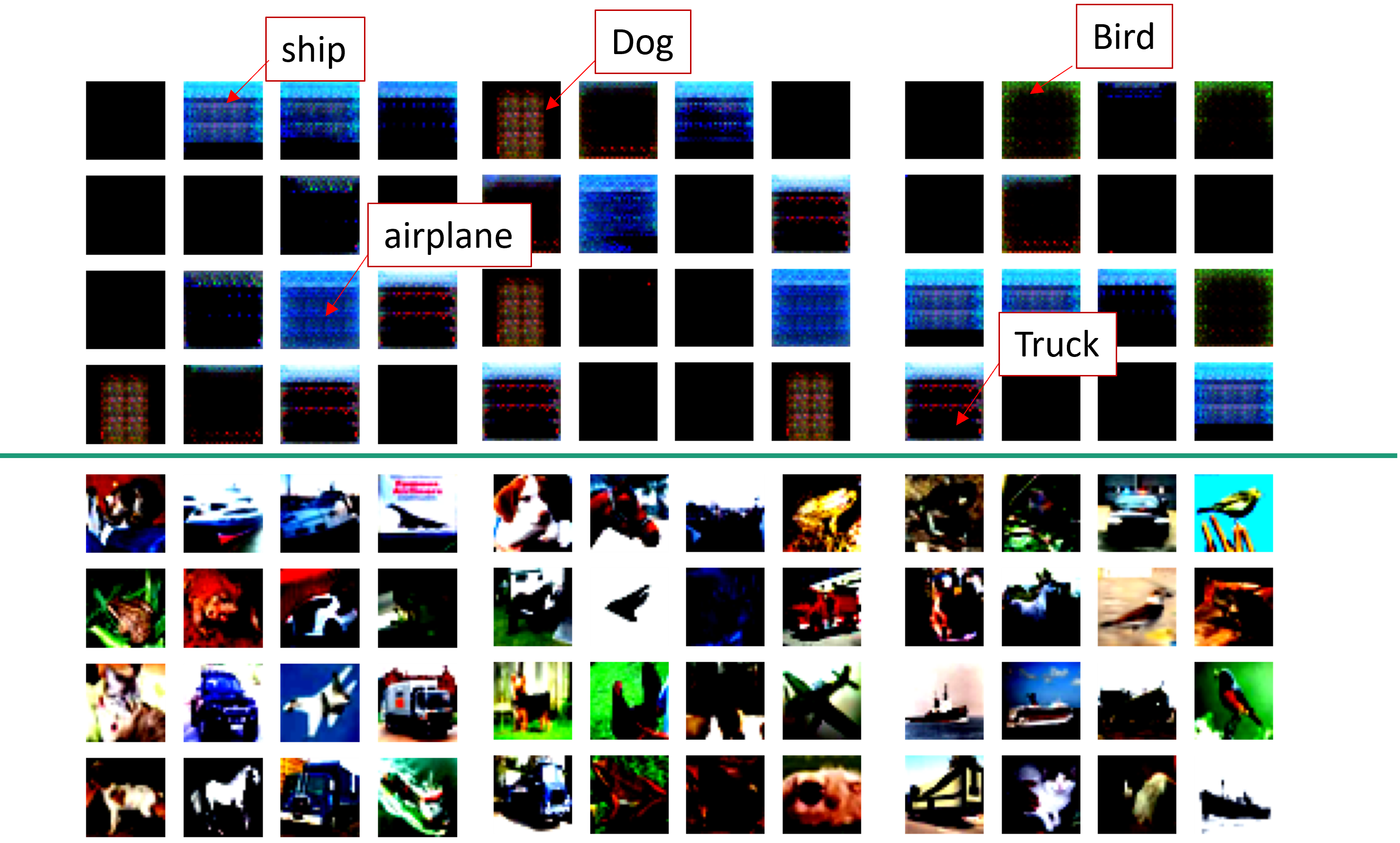}
    }
\caption{Examples of ${\cal P}_R^{-1}(D_R)$ and the corresponding raw samples $R$ (below the line) in the CIFAR-10 experiment with the deeper ${\cal P}_R$.}
\label{fig:inv-cifar10}	
\end{figure}

We further set the $SpD$ value to $1$ to simulate a relatively vulnerable scenario on the CIFAR-10 experiment.
Figure~\ref{fig:inv-cifar10} shows the results. In this scenario, we cannot successfully obtain a suitable ${\cal P}_R^{-1}$ that can recover meaningful results from a highly refined feature because of the information loss caused by the deep model structure.
Interestingly, the recovered results show different patterns that can be used to distinguish different classes. This result also suggests the effectiveness of adopting a deeper digest producer. Overall, recovering raw data from digests is difficult due to the permanent information loss caused by the feature perturbation and the steps passing through deep non-linear layers.



\section{Communication Cost and Training Time}
The communication cost and the training time are important performance indicators (aside from model accuracy) for Federated Learning. Table~\ref{tab:tc_table} shows the communication cost and training time per iteration in our EMNIST experiment.
The FedDig communication cost depends on the model gradient size and the digest size. We report both sizes in the $4$-client EMNIST experiment with $SpD=4$. For each training iteration, each client pushes the model gradient to the moderator. The gradient size of FedAvg, FedProx, and FedNova is $4.77$ MB, while the size for FedDig is $5.77$ MB due to the additional model to process the guidance. Each digest is an $8\times8\times4$ \texttt{float32} tensor. In the experiment, we transmit $139,586$ digests, which costs $177.54$ MB ($142.93$ MB for $D_R$ and $34.61$ MB for $D_y$). Note that the digests only need to be transmitted once and the size can be further reduced by compression. For the training time, the FedDig training time is slightly longer than the baselines (only less than 2x long), due to the additional training steps taken at the moderator. 
\begin{table}[t]
\caption{Communication cost and average training time per iteration on the EMNIST experiments. Values in parentheses indicate standard deviations. $n$ is the number of FL clients.}
\label{tab:tc_table}
\centerline{
\footnotesize
\begin{tabular}{lrr}
\hline
        &  \multicolumn{1}{l}{Comm. cost (MB)} &  \multicolumn{1}{l}{Training time (s)}\\ \hline
FedAvg  & $n\times 4.77$  & 24.0 (0.11) \\ \hline
FedProx & $n\times 4.77$  & 26.5 (0.07) \\ \hline
FedNova & $n\times 4.77$  & 24.3 (0.03) \\ \hline
FedDig  & $n\times 5.77+177.54$ & 44.9 (0.61) \\ \hline
\end{tabular}
}
\end{table}   

\section{Conclusion}

We present FedDig as a solution for handling data absence in Federated Learning, particularly in non-IID cross-silo scenarios. The core idea involves synthesizing model updates for missing data using a privacy-controllable data digest. These digests capture shareable information from the raw data, providing valuable guidance for training. Our evaluation includes an assessment of the digest's privacy level, and we tested FedDig in four data absence scenarios across four open datasets. The results consistently show that FedDig outperforms five state-of-the-art methods. Additionally, we thoroughly explored FedDig's hyperparameters and demonstrated its potential to integrate with existing FL algorithms.

FedDig is specifically designed for image classification, but we believe it can be adapted for other machine learning problems with reasonable modifications. The FedDig framework allows FL users to control their level of privacy, recognizing that privacy concerns are subjective and vary depending on the problem and user. Establishing clear thresholds to help users determine appropriate privacy assessment values is an intriguing area for future research, and we encourage further exploration in this direction. Additionally, since the server generates model updates for absent clients in FedDig, investigating ways to reduce the FL server's workload is another potential area for future work.

\bibliographystyle{ieeetr}
\bibliography{main}

\end{document}